%% file: main.tex
\documentclass[letterpaper]{article} 
\usepackage{aaai24}  
\usepackage{times}  
\usepackage{helvet}  
\usepackage{courier}  
\usepackage[hyphens]{url}  
\usepackage{graphicx} 
\usepackage{natbib}  
\usepackage{caption} 
\usepackage{algorithm}
\usepackage{algorithmic}

\usepackage{xcolor}
\usepackage{amsmath}
\usepackage{amsfonts}       

\usepackage{newfloat}
\usepackage{listings}
\DeclareCaptionStyle{ruled}{labelfont=normalfont,labelsep=colon,strut=off} 
\floatstyle{ruled}
\newfloat{listing}{tb}{lst}{}
\floatname{listing}{Listing}
\title{Risk-Aware Continuous Control with Neural Contextual Bandits}

\author{
    Jose A. Ayala-Romero\textsuperscript{\rm 1}
    Andres Garcia-Saavedra\textsuperscript{\rm 1}
    Xavier Costa-Perez\textsuperscript{\rm 2,\rm 1}
}
\affiliations{
    \textsuperscript{\rm 1}NEC Laboratories Europe\\
    \textsuperscript{\rm 2}i2CAT Fundation and ICREA\\
    \{jose.ayala, andres.garcia.saavedra, xavier.costa\}@neclab.eu
}

\newcommand{\name}[1]{{RANCB}}

\begin{document}

\maketitle

\begin{abstract}
Recent advances in learning techniques have garnered attention for their applicability to a diverse range of real-world sequential decision-making problems.
Yet, many practical applications have critical constraints for operation in real environments. 
Most learning solutions often neglect the risk of failing to meet these constraints, hindering their implementation in real-world contexts.
In this paper, we propose a risk-aware decision-making framework for contextual bandit problems, accommodating constraints and continuous action spaces.
Our approach employs an actor multi-critic architecture, with each critic characterizing the distribution of performance and constraint metrics.
Our framework is designed to cater to various risk levels, effectively balancing constraint satisfaction against performance.
To demonstrate the effectiveness of our approach, we first compare it against state-of-the-art baseline methods in a synthetic environment, highlighting the impact of intrinsic environmental noise across different risk configurations. 
Finally, we evaluate our framework in a real-world use case involving a 5G mobile network where only our approach consistently satisfies the system constraint (a signal processing reliability target) with a small performance toll (8.5\% increase in power consumption).
\end{abstract}

\section{Introduction}

Recent progress in the domain of decision-making learning techniques has garnered considerable attention owing to their extensive applicability in diverse real-world sequential decision-making problems \cite{go, pluribus, cicero}. Nevertheless, the practical deployment of these techniques necessitates careful consideration of critical operational constraints inherent in real environments. Regrettably, existing learning solutions often overlook the risk associated with violating these constraints, thereby impeding their viability in real-world scenarios.

Motivated by many real-world applications, we address the contextual bandit (CB) problem with constraints, which has been applied to many different problems in diverse fields, e.g., industrial control and temperature tunning \cite{pid_control}, parameter optimization in robotics \cite{safeopt}, mobile networks optimization \cite{vrain}, or video analytics optimization \cite{apostolos}.
In this framework, one metric needs to be maximized, while one or more other metrics must be bounded at each time step (step-wise constraints). 
In practice, performance metrics --- whether utility or constraints --- often possess random components. These can arise from measurement errors or be intrinsic to the metric, a phenomenon called \emph{aleatoric uncertainty}. 
Such uncertainty hinders constraint satisfaction, which is a crucial aspect in most applications, further complicating the problem.

Previous works address the aforementioned constrained contextual bandit problem considering long-term budget constraints \cite{knapsack-cb-1, knapsack-cb-2}, which does not fit our setting where the constraints must be satisfied at each step.
Other works propose linear contextual bandits with safety constraints \cite{lin-cb-constr-1, lin-cb-constr-2, TS-constr}. These solutions aim to achieve at least a percentage of the performance of a baseline policy. However, none of these works consider aleatoric uncertainty, which is a key aspect to design risk-aware decision-making algorithms.
\citeauthor{safeopt} \shortcite{safeopt} propose a Bayesian optimization algorithm called SafeOPT that handles noisy observation and constraints at each step as we do. Although SafeOPT is data-efficient, it presents important disadvantages over our solution concerning its computational complexity and requirements on prior knowledge, aspects that we discuss in detail later.

In this paper, we present a novel algorithmic framework for risk-aware decision-making.
In particular, we propose an actor multi-critic architecture. We use different critics to separately characterize the distribution of each of the metrics --- both utility and constraints.
We use these critics to train a deterministic actor that enables our solution to operate in continuous action spaces.
Previous works adopt the strategy of learning the mean value of the metric of interest \cite{dqn, td3, neuralucb}.
Other works consider a unique utility function capturing the reward with a Lagrangian-like penalty term \cite{constrained-reward-1,constrained-reward-2}.
However, such strategies can lead to constraint violations that depend on the aleatoric uncertainty inherent in the metrics. 
In contrast, our approach seeks to characterize the aleatoric uncertainty for each performance metric, which allows us to \emph{modulate the risk level in the decision-making process}.
To this end, we introduce a parameter $\alpha$ that balances between risk and performance.

We evaluate our solution against the most relevant baselines in the literature across two distinct environments. 
Firstly, in a synthetic environment where the performance metrics are non-linear functions and the set of actions that meets the constraints is highly dependent on the context. 
Within this environment, we assess the impact of aleatoric uncertainty on algorithmic performance.
Secondly, we evaluate our framework in a real-world 5G mobile network experimental platform. The primary goal here is to minimize energy consumption subject to specific system performance requirements. In this setting, we experimentally characterize the aleatoric uncertainty inherent in the system metrics.
Our solution not only shows superior constraint satisfaction but also exhibits the capability to modulate risk, effectively balancing performance against constraint satisfaction.

\section{Problem Formulation}

We consider a contextual bandit formulation with constraints. At each time step $t = 1,\ldots, T$  the learner observes the context $s_t \in \mathcal{S}$, where $\mathcal{S}$ is the context space and then selects a $d$-dimensional continuous action $a_t \in \mathbb{R}^d$. Based on this, the learner observes the reward $r_t(s_t, a_t)$ and $M$ constraint metrics $c_t^{(m)}(s_t, a_t)$, for $m = 1, \ldots, M$. All the $M+1$ observed metrics are intrinsically random, i.e., they can be written as $c_t^{(m)}(s_t, a_t) = E[c_t^{(m)}(s_t, a_t)] + \zeta_t$, where the noise $\zeta_t$ at time $t$ is drawn from an unknown distribution with expectation $E[\zeta_t] = 0$.
The behavior of the learner is defined by a policy that maps contexts into actions $\pi : \mathcal{S} \mapsto \mathbb{R}^d$. Our objective is to find the optimal policy:
\begin{align}\label{eq:main_prob}
  & \operatorname*{argmax}_{\pi} \sum_{t=1}^T r_t(s_t, \pi(s_t))  & \\
  & \textup{s.t.} \quad c_t^{(m)}(s_t, \pi(s_t)) < c^{(m)}_{\text{max}}, & m &= 1, \ldots, M \nonumber\\ 
  & & t &= 1, \ldots, T \nonumber 
\end{align}
where $c^{(m)}_{\text{max}}$ is the maximum value for constraint $m$.

Note that, in contrast to the problem addressed in other works on contextual bandits in the literature \cite{neuralucb,neurallinucb,lin-cb-constr-1, lin-cb-constr-2}, we consider a continuous action space, several performance metrics, and stochastic constraints that should be satisfied at each time step.

\section{Proposed Method}

We consider an actor-multi-critic architecture with a deterministic actor to deal with the continuous action space \cite{ddpg}.
In contrast to previous works, we consider $M+1$ distributional critics denoted by $R^m(s,a\mid \eta^m)$, where $\eta^m$ are the parameters of the critics. Note that the critic with index $m=0$ approximates the reward function $r_t(s_t, a_t)$ and the critics with indexes $m=1, \ldots, M$ approximate constraint $c_t^{(m)}(s_t, a_t)$.
We enable the critics to approximate the distribution of their objective metric using quantile regression.

\subsection{Distributional Critics} \label{sec:secccc}

Let $F_Z(z)$ be the cumulative distribution function (CDF) of $Z$. Note that the quantile function is the inverse of the CDF. Hence, for a given quantile $\tau \in [0, 1]$, the value of the quantile function is defined as $q_\tau = F^{-1}_Z(\tau)$.  
The quantile regression loss is an asymmetric convex function that penalizes overestimation error with weight $\tau$ and underestimation error with weight $1-\tau$:
 \begin{align}\label{eq:qr_loss}
    \mathcal{L}^\tau (\hat{q}_\tau) &:= \mathbb{E}_{z\sim Z} \left[ \rho_\tau (z - \hat{q}_\tau) \right], \text{where} \\
    \rho_\tau (u) &:= u\cdot (\tau - \delta_{\{u<0\}}) \;\;\;\; \forall u \in \mathbb{R},
\end{align}
where $\hat{q}_\tau$ is the estimation of the value of the quantile function, and $\delta_{\{x\}}$ is an indicator function that takes the value $1$ when the condition $x$ is satisfied and $0$ otherwise.
Considering that each critic has $N$ outputs that approximate the set $\{ q_{\tau_1}, \ldots, q_{\tau_N} \}$, the critic can be trained to minimize the following objective using stochastic gradient descent:
\begin{equation}\label{eq:loss_all}
    \sum_{i=1}^N \mathcal{L}^{\tau_i} (\hat{q}_{\tau_i}).
\end{equation}

Note that the quantile regression loss is not smooth when $u=0$, limiting the performance of non-linear function approximators such as NNs. To address this issue, we use the \textit{quantile Huber loss}~\cite{huber}. This loss function has a squared shape in an interval $[-\kappa, \kappa]$, and reverts to the standard quantile loss outside of this interval:
\begin{equation}
 L_\kappa (u) :=
  \begin{cases}
    \frac{1}{2} u^2       & \quad \text{if } |u| \leq \kappa \\
    \kappa (|u| - \frac{1}{2} \kappa )  & \quad \text{otherwise.} 
  \end{cases}
\end{equation}

Now, we derive an asymmetric variation of the Huber loss,
\begin{equation}\label{eq:huber}
    \rho^\kappa_\tau (u) := |\tau - \delta_{\{u<0\}}| \frac{L_\kappa (u)}{\kappa}.
\end{equation}

Finally, the quantile Huber loss can be obtained by introducing $\rho^\kappa_\tau (u)$ in eq~\eqref{eq:qr_loss}. Note that when $\kappa \to 0$ the quantile Huber loss reverts to the quantile regression loss.

\subsection{Risk-Aware Actor}

In order to capture the information provided by all the critics, we define an aggregate reward signal:
\begin{align}\label{eq:ragg}
    & R^{agg}(s,a,\alpha \mid \eta)  := \\
    & \bar{R}^0(s,a \mid \eta^0) - \sum_{i=1}^{M} \lambda \max\left( \gamma^{\alpha}(R^i(s,a \mid \eta^i)) - c^{(i)}_{\text{max}} , 0 \right), \nonumber
\end{align}
where $\gamma^{\alpha}(Z)$ is the value of the quantile function of distribution $Z$ at quantile $\alpha$, $\lambda$ is the penalty constant for the constraints, $\eta = \{\eta^0, \ldots, \eta^M\}$ is the joint set of parameters of the $M+1$ critics, and $\bar{R}^m(\cdot)$ indicates the mean of the distribution provided by critic $m$.
That is, the first term in eq.~\eqref{eq:ragg} is the mean of the metric we want to maximize, while the second term captures the penalty incurred when violating each constraint. Note that, when $\gamma^{\alpha}(R^i(s,a \mid \eta^i)) < c^{(i)}_{\text{max}}$, the value inside the $\max()$ function is negative and the penalty terms are zero.

Importantly, in contrast to other works using similar weighted penalties in the reward, the use of the quantile function allows us to assure that the tail of the distribution of the constraints (as $\alpha \rightarrow 1$) meets the restrictions, making our solution more robust to constraint violations. 
Note that eq.~\eqref{eq:ragg} can be adapted to cases where the constraints set a minimum value by changing the sign to the term inside the maximum operator and choosing a value of $\alpha$ close to zero.

\begin{figure}[t!] 
\centering
\includegraphics[width=0.85\linewidth]{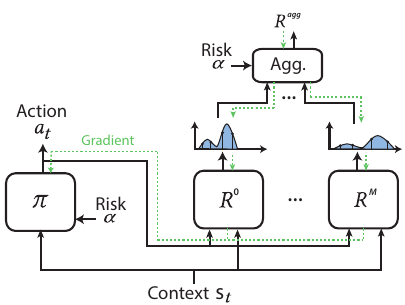}
\caption{\small{Risk aware decision-making framework comprising a deterministic actor, $M+1$ distributional critics and the aggregation function detailed in eq.~\eqref{eq:ragg}. The propagation of the gradient to train the actor is shown in green. }}
\label{fig:model}
\end{figure}

We denote the deterministic actor policy as $\pi(s \mid \theta, \alpha)$, where $\theta$ is the set of actor parameters.
Then, for a given value of $\alpha$, we define the actor's  objective as
\begin{align}\label{eq:J}
    J(\pi, \alpha) :&=  \int_\mathcal{S} \beta(s) \, R^{agg}(s, \pi(s \mid \theta, \alpha),\alpha \mid \eta)\,\mathrm{d}s  \\
    & = \mathbb{E}_{\mathbf{s} \sim \beta} [R^{agg}(s, \pi(s \mid \theta, \alpha),\alpha \mid \eta)],  \nonumber
\end{align}
where $\beta(s)$ is the stationary context distribution. Note that, in a contextual bandit problem, the distribution of the context is not conditioned by the policy. 

The actor policy is updated by applying the chain rule to the performance objective defined in eq.~\eqref{eq:J} with respect to the actor parameters \cite{dpg}:
\begin{align}\label{eq:policy_gradient}
 \nabla_{\theta} & J(\pi, \alpha) \approx \\
   &\mathbb{E}_{s \sim \beta} \left[ \nabla_{a} R^{agg}(s, a,\alpha \mid \eta)\mid_{a=\pi (s \mid \theta, \alpha)} \nabla_{\theta} \pi (s \mid \theta, \alpha) \right]. \nonumber
\end{align}

Note that $\alpha$ is an input of the policy and the aggregated reward (eq.~\ref{eq:ragg}). Thus, different values of $\alpha$ can modulate the risk taken by the actor when selecting actions. Specifically, with $\alpha \to 1$  we reduce the probability of violating a constraint. However, this may also imply lower values of reward $r_t$ due to more conservative actions, showing the trade-off between performance and robustness.
Moreover, we can configure diverse values of $\alpha$ for each constraint when the constraints have different risk aversion (e.g., some constraints may be more critical than others). This is possible because we consider one critic per constraint that can learn with a different value of $\alpha$. Then, $R^{agg}(\cdot)$ is computed based on the corresponding values of $\alpha$ from each critic.
The proposed framework for risk-aware decision-making is shown in Fig.~\ref{fig:model}.

\subsection{Algorithm}

We consider a reply buffer $\mathcal{D}$, where the samples of experience $\langle s, a, c^{(0)}(s,a), \ldots,  c^{(M)}(s,a)\rangle$ are stored at each time step \cite{dqn}. To simplify the notation, the reward $r(s,a)$ is denoted by $c^{(0)}(s,a)$ in the reply buffer.
The gradients used for training are computed using mini-batches of $B$ experience samples randomly gathered from this buffer.

Let $\alpha$ be the default risk value. In most of the applications, a risk-averse policy is desirable, i.e., $\alpha \to 1$.
In other cases, we may want to modulate the level of risk to find a different balance between performance and robustness. We define $\mathcal{A}$ as the set of risk values to be used by the algorithm.
We define $\mathcal{T}$ as the set of quantiles approximated by all the critics, where $\mathcal{A} \subseteq \mathcal{T}$.
For any given minibatch of experience samples, $R^{agg}$ can be computed for different values of risk $\alpha$ (see eq.~\eqref{eq:ragg}). Thus, the actor can learn the policy as a function of the risk without collecting extra data.
Since the actor policy $\pi$ is deterministic, we add some noise denoted by $\mathcal{N}$ to the actions to enable exploration during training. 
Algorithm~\ref{alg:algorithm} shows the pseudo-code of our framework, referred to as Risk-Aware Neural Contextual Bandit (\name{}).

\begin{algorithm}[tb]
    \caption{\name{} training}
    \label{alg:algorithm}
    \textbf{Input}: $B$, $\mathcal{A}$, $\alpha$, $\kappa$ $\mathcal{T}$\\
    \textbf{Initialize}: $\mathcal{D} = \emptyset$, $\mathcal{N}$, $\theta$, $\eta$
    
    \begin{algorithmic}[1] 
        \FOR{$t = 1, \ldots, T$}
            \STATE Observe context $s_t$
            \STATE Compute the action $a_t = \pi(s_t, \alpha \mid \theta) + \mathcal{N}_t$
            \STATE Observe performance metrics $r_t, c^{(1)}_t, \ldots c^{(M)}_t$
            \STATE Store in $\mathcal{D}$ the experience $\langle s_t, a_t, c_t^{(0)} \ldots  c_t^{(M)}\rangle$
            \STATE Sample a random minibatch of $B$ samples $\langle s_i, a_i, c_i^{(0)} \ldots  c_i^{(M)}\rangle$
            \FOR{ $m = 0, \ldots, M$}
                \STATE Update the critic $m$ by minimizing the loss \\ $L = \frac{1}{B} \sum_i \sum_{\tau \in \mathcal{T}} \rho^\kappa_{\tau} (c^{(m)}_i - \gamma^{\tau}(R^m(s_i,a_i | \eta^m)))$
            \ENDFOR
            \FOR{ all $\alpha_j \in \mathcal{A}$}
                \STATE Update the actor using the sampled policy gradient \\ $\frac{1}{B} \sum_i \nabla_{a} R^{agg}(s_i, a,\alpha_j | \eta)|_{a=\pi (s_i | \theta, \alpha_j)} \nabla_{\theta} \pi (s_i | \theta, \alpha_j)$ 
            \ENDFOR
        \ENDFOR
    \end{algorithmic}
\end{algorithm}

\section{Benchmark Algorithms}

We first present a set of benchmarks that are variations of our proposal and are inspired by ideas from the literature. In this way, we can conduct an ablation study to evaluate the impact of its different components, i.e., distributional critics and multiple critics.
Then, we present SafeOPT \cite{safeopt}, the most closely related work to ours. SafeOPT relies on GPs to learn the objective and the constraint functions while handling the intrinsic noise of the observations. To the best of our knowledge, there are no other works in the literature addressing this problem.

\subsection{Baselines}

\begin{itemize}
    \item \textbf{Neural Contextual Bandit (NCB)} is inspired by the actor-critic NN architecture presented by \citeauthor{ddpg} \shortcite{ddpg}. However, some modifications need to be introduced to adapt this solution to our problem. As NCB only encompasses one critic, we need to define a utility function that captures the constrained problem:
    \begin{align}\label{eq:utility_ncb}
     u_t & (s_t, a_t) := \\
       & r_t(s_t, a_t) - \sum_{i=1}^{M} \lambda \max\left( c_t^{(m)}(s_t, a_t) - c^{(i)}_{\text{max}}, 0 \right). \nonumber
    \end{align}
    In contrast to the original algorithm \cite{ddpg} where future values of reward are also taken into account, the NCB critic approximates the expectation of $u_t$. For that purpose, we use the MSE loss function: 
    \begin{equation}\label{eq:ncb_critic_loss}
    L(\eta) := \mathbb{E}_{s \sim \beta, a \sim \pi'} \left[ \left( R(s,a|\eta) - u(s,a) \right)^2 \right]
    \end{equation}
    where $R(s,a|\eta)$ denotes the critic and $\pi'$ any policy that can potentially deviate from the actor's behavior. The actor is updated as indicated by \citeauthor{ddpg} \shortcite{ddpg}.
    
    \item \textbf{Single-Critic Distributional NCB (SC-DNCB)} extends NCB with a distributional critic to characterize the distribution of $u_t$. 
    The critic is trained using the Huber loss in eq.~\eqref{eq:huber} and the actor uses the policy gradient equation proposed by  \citeauthor{ddpg} \shortcite{ddpg} with respect to the expectation of the distribution provided by the critic.
    Note that, as the reward and all the constraints are characterized by a single utility function, the level of risk tolerance in the decision-making process cannot be configured. However, it has been widely reported in the literature that the use of distributional critics increases the performance of the algorithms even when the critic is only used to compute the expected value of the distribution \cite{c51,qr-dqn}. 
    
    \item \textbf{Multi-Critic NCB (MC-NCB)} extends NCB by including $M+1$ non-distributional critics, one per performance metric. Each critic approximates the expectation of its corresponding metric using the MSE loss. To update the actor, an aggregated reward signal is computed based on the output of all the critics, similarly to eq.~\eqref{eq:ragg}. Then, the actor gradients are computed using eq.~\eqref{eq:policy_gradient}.  
\end{itemize}

All these baseline solutions use the same exploration approach used by  \name{} in the training phase. 
Note that none of these baselines allow us to configure the level of risk during the decision-making process as \name{} does.

\subsection{Bayesian Optimization with Constraints}
Finally, we also use SafeOPT as a benchmark \cite{safeopt}, which is a Bayesian online learning algorithm that handles constraints and noisy observations. 
%
SafeOPT comprises $M+1$ GPs that characterize each one of the performance and constraint metrics. We implement the contextual version of SafeOPT and follow the implementation provided in Sec.~4.3 of that paper, where the confidence intervals of the GPs are used to compute the safe set of actions (actions that satisfy the constraint for a given context). The confidence is determined by a scalar $\beta$ (see eq.~(10) in \cite{safeopt}). 
We consider this practical version of the algorithm because we assume that, in general, the Lipschitz continuity properties of the performance and constraint functions are unknown.

We configure SafeOPT with a combination of the anisotropic version of the Matér kernel with $\nu = \frac{3}{2}$ and a white kernel to model the noise \cite{kernels}.
We use UCB as an acquisition function since it optimizes the reward and expands the safe set of actions at the same time. 
We found that this strategy provides higher performance compared to the exploration strategy proposed originally by SafeOPT, which expands the safe set explicitly. We also note that this issue has been reported in other previous works \cite{pid_control, edgebol_tmc}.

During the execution of SafeOPT, it may happen that none of the actions satisfies the conditions to be in the safe set, e.g., due to a large value of $\beta$ or high noise in the observations. The original algorithm does not consider that this event may happen in practice. To address this issue, we modified SafeOPT to select the action that minimizes the estimation of the accumulated constraint violation (across all the constraints) when the safe set is empty.

\section{Evaluation}
We evaluate all the aforementioned algorithms in two settings: ($i$) using a synthetic environment with non-linear functions and variable noise in the observations, and ($ii$) in a real-world resource allocation problem in wireless networks implemented on a real system.
The source code of our solution and all the baselines is available online\footnote{\url{https://github.com/jaayala/risk_aware_contextual_bandit}}.

In our evaluation, we configure all actor and critic NNs with two hidden layers of 256 units. 
The critics that approximate the reward function learn the set of quantiles $\mathcal{T} = \{i/N \mid i=1, \ldots, N \}$, where $N=21$.
For the critics that approximate constraints, we consider two different configurations depending on whether the constraint sets a maximum value (synthetic environment) or a minimum value (resource allocation problem in wireless networks). For the former, we use $\mathcal{T}^{\text{max}} = \{0.1, 0.3, 0.5, 0.7, 0.8, 0.9, 0.99, 0.995, 0.999\}$ and $\alpha^{\text{max}} = 0.995$; and for the latter $\mathcal{T}^{\text{min}} = \{1-\tau \mid \tau \in \mathcal{T}^{\text{max}}\}$ and $\alpha^{\text{min}} = 0.005$.
We used Adam to learn the NN parameters with a learning rate of $10^{-4}$ and $10^{-3}$ for the actor and critics, respectively.
For the exploration noise $\mathcal{N}$, we use an Ornstein-Uhlenbeck process with the parameters $\theta_{\text{noise}} = 0.15$ and $\sigma_{\text{noise}} = 0.15$ to generate temporally correlated perturbations to the selected action \cite{ddpg}.
We use a reply buffer $\mathcal{D}$ with a memory of 2000 samples.
Finally, we configure $\kappa=1$, a minibatch size of $B = 64$ samples, and $\lambda = 2.5$ (see the Appendix for a detailed evaluation).

For all the results shown in this section, we consider 10 independent runs. The figures with shadowed area show the average and the 15$^{\text{th}}$ and 85$^{\text{th}}$ percentiles. The figures with error bars show the mean values and the confidence intervals with a confidence level of 0.95.

\subsection{Synthetic Environment}
\begin{figure}[t!] 
\centering
\includegraphics[width=0.99\linewidth]{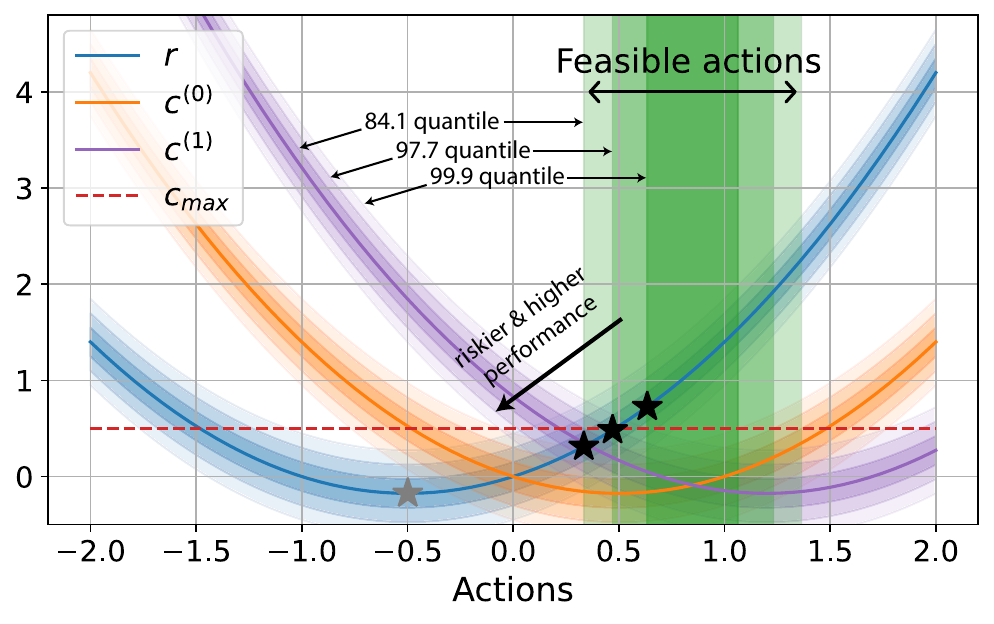}
\caption{\small{Representation of the synthetic environment defined in eq.~\eqref{eq:syn_env} for a fixed context $s=(0.7, 0.7, 0.7)$ and $\sigma_{\text{env}} = 0.15$. We depict the 84.1$^{\text{th}}$, 97.7$^{\text{th}}$, and 99.9$^{\text{th}}$ quantiles of the functions with different transparency levels and their corresponding sets of feasible actions. The markers show the optimal values of the unconstrained (grey) and constrained (black) problems.
}}
\label{fig:syn_env}
\end{figure}

In the first set of experiments, we consider a synthetic environment with 3-dimensional contexts and a one-dimensional action space ($d=1$). The reward and the constraints are given by the following quadratic functions:
\begin{gather}\label{eq:syn_env}
 r_t(s_t,a_t) = s_t^{(0)} \cdot a_t^2 + s_t^{(1)} \cdot a_t  + \xi_t^{(0)}\\
 c_t^{(1)}(s_t,a_t) = s_t^{(0)} \cdot a_t^2 - s_t^{(1)} \cdot a_t  + \xi_t^{(1)}, \nonumber \\ 
 c_t^{(2)}(s_t,a_t) = s_t^{(0)} \cdot (a_t - s_t^{(2)})^2 - s_t^{(1)} \cdot (a_t  - s_t^{(2)})  + \xi_t^{(2)}, \nonumber
 \end{gather}
where $\xi_t^{(i)} \sim N(0,\sigma_{\text{env}}^2)$ for $i=0,1,2$. In our experiments, the contexts are generated as i.i.d. uniform random variables in $[0, 1]^3$. We set the constraint bounds as $c_{\text{max}}^{(0)} = c_{\text{max}}^{(1)} = 0.3$. 

Fig.~\ref{fig:syn_env} shows an example of the functions in eq.~\eqref{eq:syn_env} for a fixed context $s = (0.7, 0.7, 0.7)$ and $\sigma_{\text{env}} = 0.15$. In this example, the lower value of the feasible action set is outlined by $c_t^{(2)}$ and the higher value is delimited by $c_t^{(1)}$. Note that the location and shape of all of these functions are highly dependent on the context $s$.

We plot with different transparencies the  84.1$^{\text{th}}$, 97.7$^{\text{th}}$, and 99.9$^{\text{th}}$ quantiles of the functions in eq.~\eqref{eq:syn_env}. We also plot the feasible sets of actions obtained when considering that each of those quantiles needs to satisfy the constraint. Note that when considering higher quantiles, the safe set becomes smaller but safer, i.e., the probability of satisfying the constraints is higher and vice versa.
The optimal values of the reward $r$ for each feasible set of actions are marked with a black star, and we use a grey star for the optimal unconstrained value. 
Note that the riskier the set of feasible actions, the higher the performance of the optimal action within the set, i.e., we get closer to the grey star.

In other words, a higher reward implies a higher probability of violating the constraint. Therefore, if we want to satisfy the constraint with high probability, we need to be more conservative in decision-making, which has a cost in terms of reward.
This trade-off also depends on the variance of the noise of the performance metrics, modeled by $\sigma_{\text{env}}^2$.

\begin{figure}[t!] 
\centering
\begin{minipage}{\linewidth}
\centering
\includegraphics[width=0.49\linewidth]{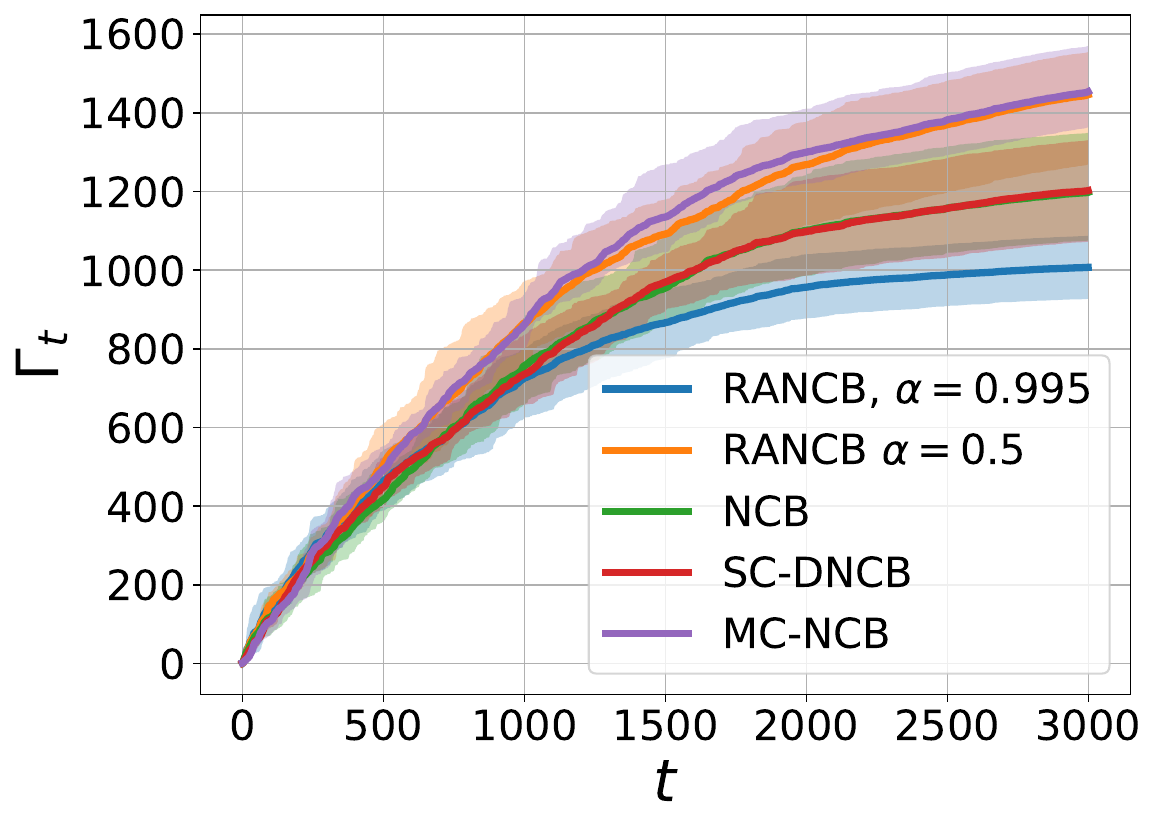}
\includegraphics[width=0.49\linewidth]{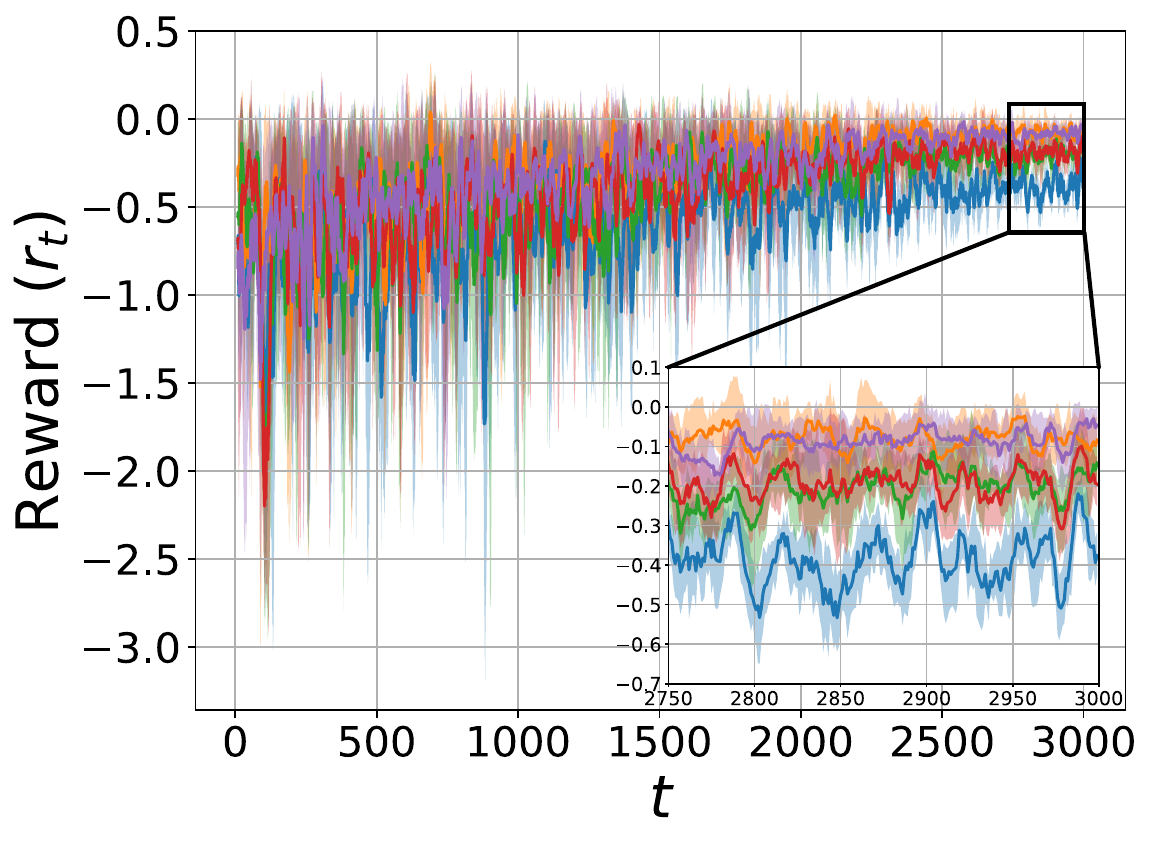}
\end{minipage}
\caption{\small{Evaluation of training phase in synthetic environment with $\sigma_{\text{env}} = 0.2$. Accumulated constraint violation $\Gamma_t$ (left); instantaneous reward $-r_t(s_t, a_t)$ (right).}}
\label{fig:syn_train}
\end{figure}

Fig.~\ref{fig:syn_train} compares the performance of all the solutions during training. The right plot shows the instantaneous reward ($-r_t(s_t, a_t)$), and the left plot shows the accumulated constraint violation defined as follows:
\begin{equation}\label{eq:constr_viol}
\Gamma_t := \sum_{m = 1}^{M} \sum_{t' = 0}^{t} \max \left\{ c_{t'}^{(m)} - c_{\text{max}}^{(i)}, 0 \right\}.
\end{equation}

We observe that \name{} with $\alpha=0.995$ not only provides the minimum values of $\Gamma_t$ but also the slope of $\Gamma_t$ tends to zero. This means that the constraint violation after convergence is very small in this setting. Obviously, this outstanding reliability performance comes at a cost in terms of reward as depicted in the right plot. 
Conversely, \name{} with $\alpha=0.5$ provides the highest reward but pays the price of higher accumulated constraint violations.
This result illustrates how \name{} can adapt to any application reliability target by setting $\alpha$ appropriately. 
The rest of the benchmarks are unable to adjust the level of risk and, therefore, they converge to intermediate solutions.

Note that Fig.~\ref{fig:syn_train} does not include SafeOPT.  
The reason is that it is not feasible to provide a fair comparison of the training performance between SafeOPT and the rest of the algorithms due to some fundamental differences.
On the one hand, differently than our approach, SafeOPT needs some previous knowledge before starting the training phase.
First, SafeOPT needs a dataset to optimize the hyperparameters of the kernels. As the kernels encode the smoothness of the metric function, this step is critical. A suboptimal hyperparameter optimization (e.g., due to a poor dataset) may have serious consequences on training performance. 
In this particular example, there are 5 hyperparameters to optimize for each GP, i.e., 4 dimensions (3-dimensional contexts and 1-dimensional action) plus the noise level. In our evaluations, we optimized the kernels using 1000 samples obtained randomly from the environment.
Second, we need to define an initial safe set of actions, which will be used at the beginning of the training phase. This step requires some domain knowledge and can be very challenging as the safe set of actions can be highly dependent on the context. We found that, if the actions in the initial safe set violate the constraint, the algorithm does not converge. To avoid this, we use eq.~\eqref{eq:syn_env} in the first iterations of the training phase to compute an initial safe set of actions. Note that this gives SafeOPT some advantage over the other benchmarks, hindering a fair comparison. Moreover, this strategy to generate the initial safe set is not realistic in a real-world application.

\begin{figure}[t!] 
\centering
\begin{minipage}{0.48\linewidth}
\includegraphics[width=\linewidth]{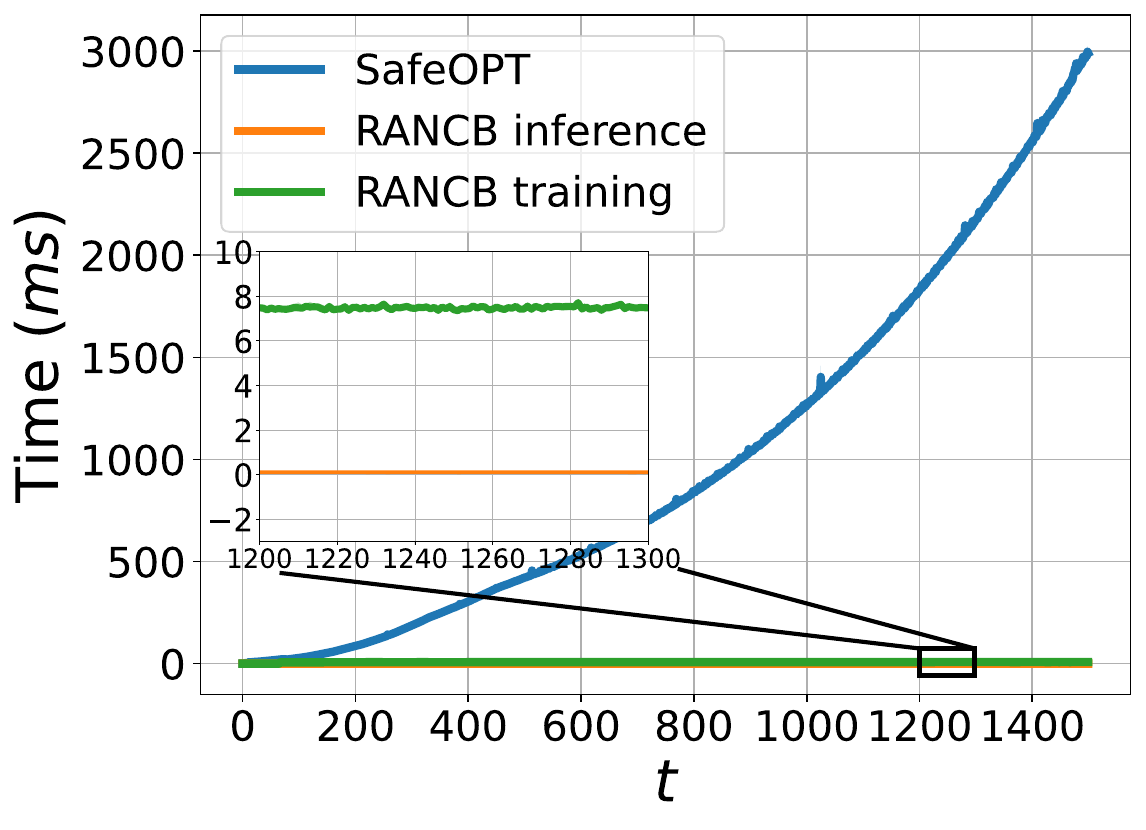}
\caption{\small{Evaluation of execution time in a Intel i7-11700 @ 2.5GHz and 15Gb or RAM.}}
\label{fig:times}
\end{minipage}
\hfill
\begin{minipage}{0.48\linewidth}
\centering
\includegraphics[width=\linewidth]{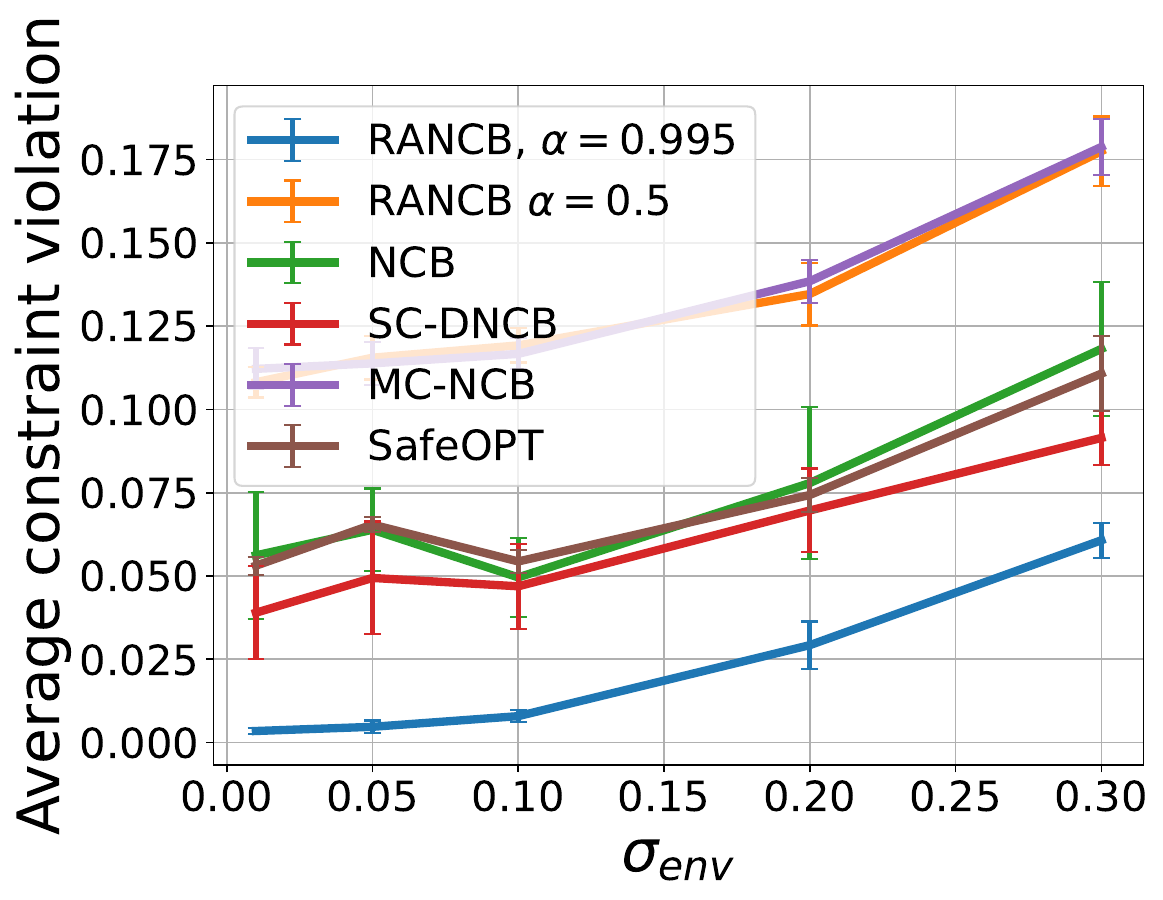}
\caption{\small{Evaluation of inference performance. Average constraint violation per step as a function of $\sigma_{\text{env}}$.}}
\label{fig:syn_eval1}
\end{minipage}
\end{figure}

On the other hand, GP-based learning algorithms are known to be more data efficient than NN-based algorithms, that is, they need fewer data samples to converge. However, the computational complexity of GP-based solutions is $O(n^3)$ with the sample size \cite{gp-book}. 
To illustrate this, Fig.~\ref{fig:times} shows the execution times of SafeOPT and \name{} in an Intel i7-11700 @ 2.5GHz with 15Gb of RAM. The execution time of SafeOPT increases exponentially with the sample size, while the inference time of \name{} is $0.106 \pm 5 \cdot 10^{-4}$~ms\footnote{Similar times are observed when using a GPU NVIDIA A100-SXM4-80GB. Due to the small size of the NNs, there is no noticeable gain in execution time when using a GPU.}. Additionally, we measured that the execution time for the hyperparameter optimization of SafeOPT is $160.7 \pm 15.1$ seconds for each GP.
Therefore, GP-based solutions can be unfeasible in scenarios where the computational capacity is limited or the decisions need to be made synchronously or in a timely manner (as in the wireless network example that we show later). 

Let us now compare the performance of all the benchmarks during the inference operation (after the training phase).
Fig.~\ref{fig:syn_eval1} shows the average constraint violation as a function of $\sigma_{\text{env}}$.
Note that, with larger values of $\sigma_{\text{env}}$, there are contexts for which the penalty term in eq.~\eqref{eq:ragg} is not zero for any action, i.e., the constraint violation is inevitable due to the high variance. Hence, we observe a general increase of the constraint violation with $\sigma_{\text{env}}$.
In such cases, the algorithms need to select the action that minimizes the cost due to constraint violation. 
In all the cases, \name{} with $\alpha=0.995$ obtains the minimum constraint violation. 
Moreover, we found that optimal values of the hyperparameter $\beta$ of SafeOPT are highly dependent on $\sigma_{\text{env}}$. 
We evaluated several values of $\beta$ and selected for each $\sigma_{\text{env}}$ the one attaining the lowest constraint violation, $\beta=\{90, 15, 10, 3.5, 2\}$ for each value in the x-axis of Fig.~\ref{fig:syn_eval1}, respectively.

\begin{figure}[t!] 
\centering
\begin{minipage}{\linewidth}
\centering
\includegraphics[width=0.49\linewidth]{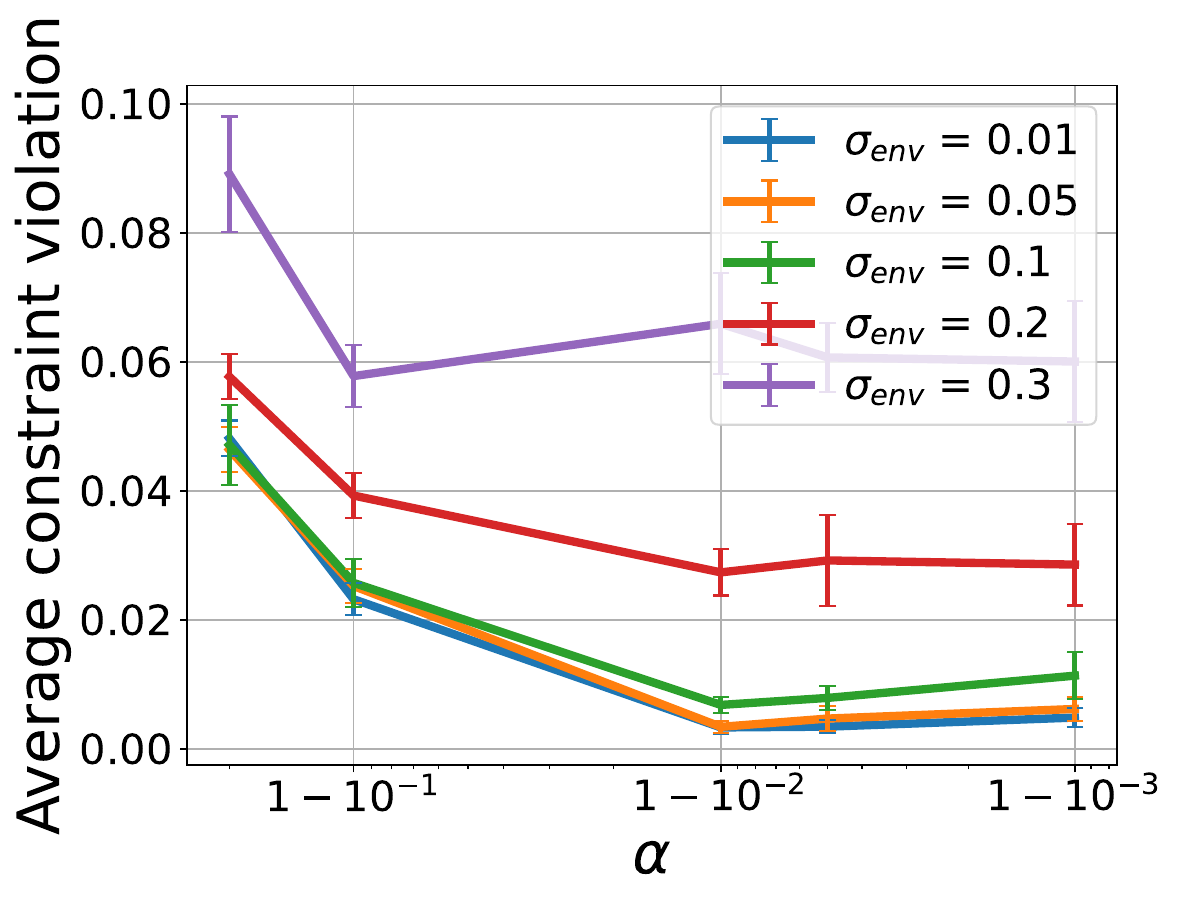}
\includegraphics[width=0.49\linewidth]{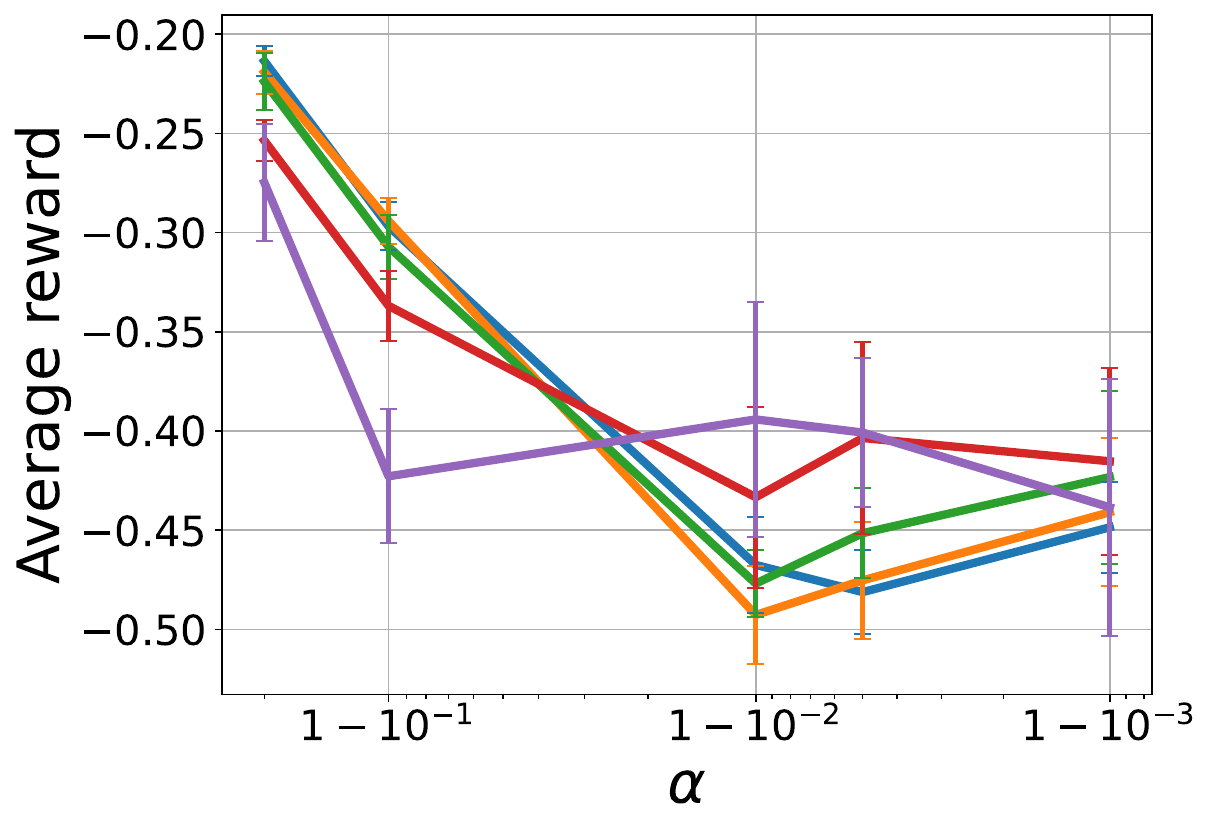}
\end{minipage}
\caption{\small{Impact $\alpha$ on the execution performance of \name{}. Average constraint violation per step (left) and average reward (right).}}
\label{fig:syn_eval2}
\end{figure}

Finally, Fig.~\ref{fig:syn_eval2} shows the impact of $\alpha$ on the inference performance of \name{} for different values of $\sigma_{\text{env}}$. 
As expected, when $\alpha$ increases, the constraint violations decrease (left plot).
We also observe that lower values of $\alpha$ are associated with higher reward (right plot), which shows again the trade-off between constraint satisfaction and performance.

\subsection{Resource Assignment in Mobile Networks}

For every Transmission Time Interval (TTI) of 1~ms or lower, wireless processors such as those in 5G must process signals that encode data, known as Transport Blocks (TB), within hard time deadlines. Failing to meet such deadlines may result in TB data loss~\cite{concordia}. 
To provide industry-grade reliability, today's wireless processors use hardware accelerators (HAs) that can swiftly process these signals.
However, it is well-known that HAs are energy-hungry, and energy consumption is nowadays a major concern for mobile operators \cite{gsma-energy-2, china-mobile-energy}.
Alternatively, software processors, which use inexpensive CPUs, are more energy-efficient but are slower than HAs, potentially risking deadline violations. 

Importantly, the processing time of these signals (in a software processor or an HA) is difficult to predict as it depends on several and potentially unknown variables, i.e., the TB size, and the signal quality, among others \cite{concordia}. 
Thus, we face a resource assignment problem where we need to decide between energy-efficient CPUs or high-performing HAs to process incoming signals with uncertain processing times. 
In this context, an overuse of CPUs to save energy may cause that many TBs are not processed within their deadlines leading to data loss, which has serious implications for the mobile operator. In other words, there is a trade-off between processing constraints (deadlines when processing signals) and energy consumption. 

\begin{figure}[t!] 
\centering
\includegraphics[width=0.95\linewidth]{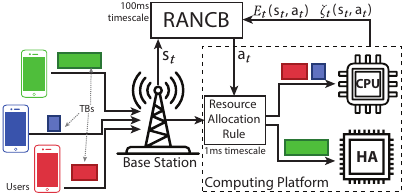}
\caption{\small{Simplified scheme of the resource assignment problem in mobile networks. 
}}
\label{fig:networking_scheme}
\end{figure}

Current Open RAN (O-RAN) systems support third-party applications for resource control at 100 millisecond timescales~\cite{oran-magazine}. This timescale brings an additional challenge since the decisions cannot be made per TB but with a coarser time granularity. As shown in Fig.~\ref{fig:networking_scheme}, our algorithm makes resource allocation decisions every 100~ms, which are implemented as \emph{rules} that are then applied to each TB in real-time in the computing platform.

We hence formulate this problem as the following constrained contextual bandit. 
We define $s_t$ as the traffic characterization at time $t$ (context). The offloading decision is denoted by $a_t \in [0, 1]$. The energy consumption of the system (in Joules) for a given context and action is denoted by $E_t(s_t, a_t)$. The ratio of TBs that have been processed within their deadline (reliability) is denoted by $\zeta_t (s_t, a_t) \in [0, 1]$.
We formulate the problem as follows:
\begin{align}\label{eq:network_problem}
  &\underset{\{a_t\}_{t=1}^{T} }{\min} &&  \lim_{T\rightarrow \infty} \frac{1}{T} \sum_{t=1}^T  E_t(s_t, a_t) & \\
  & \textup{s.t.} && \lim_{T \rightarrow \infty} \frac{1}{T} \sum_{t=1}^T \zeta_t(s_t, a_t) \ge 1-\epsilon. \nonumber
\end{align}
\noindent where $\epsilon$ sets the target reliability. More details about the formulation are provided in the Appendix.
We would like to highlight that both $E_t(\cdot)$ and $\zeta_t(\cdot)$ are very complex functions whose closed-form expressions are unavailable. They characterize the high complexity of the system (i.e., the number of users and their mobility patterns, signal quality, app data generation, etc.) during a period of 100 ms and also depend on the specific hardware of the system. For these reasons, they need to be learned from observations.

\begin{figure}[t!] 
\centering
\begin{minipage}{\linewidth}
\centering
\includegraphics[width=0.49\linewidth]{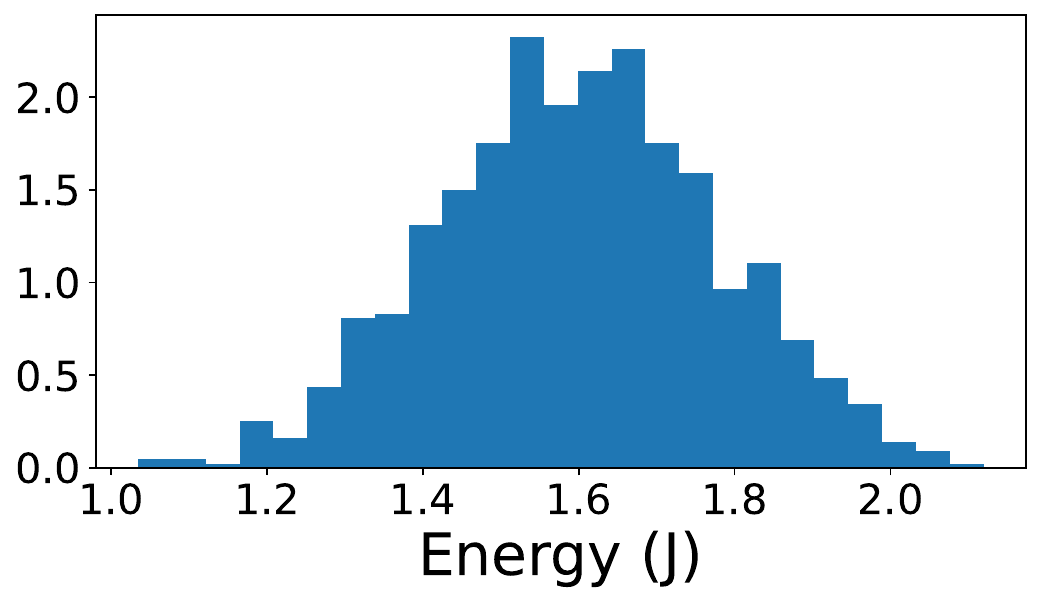}
\includegraphics[width=0.49\linewidth]{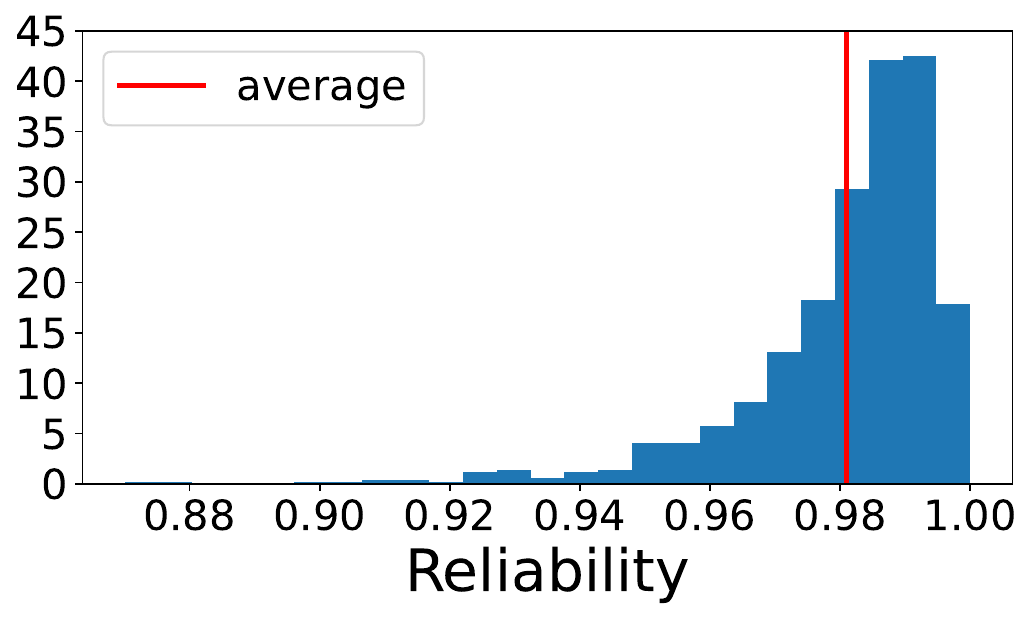}
\end{minipage}
\caption{\small{Empirical probability density function of the $E(\cdot)$ and $\zeta(\cdot)$ for a fixed decision $a = 0.6$} and stationary traffic load.}
\label{fig:network_distr}
\end{figure}

Using our experimental platform detailed in the Appendix \cite{salvat2023open}, we characterized experimentally the distribution of the energy ($E(\cdot)$) and the system's reliability ($\zeta(\cdot)$) using a fixed offloading decision $a = 0.6$.
Fig.~\ref{fig:network_distr} shows that these metrics are random in nature. In particular, if we only considered the average value of the reliability to satisfy the constraint, the reliability would be below its minimum required value with a probability of 0.35 for the distribution in Fig.~\ref{fig:network_distr}. 
Our proposal aims at minimizing this risk.

\begin{figure}[t!] 
\centering
\begin{minipage}{\linewidth}
\centering
\includegraphics[width=0.49\linewidth]{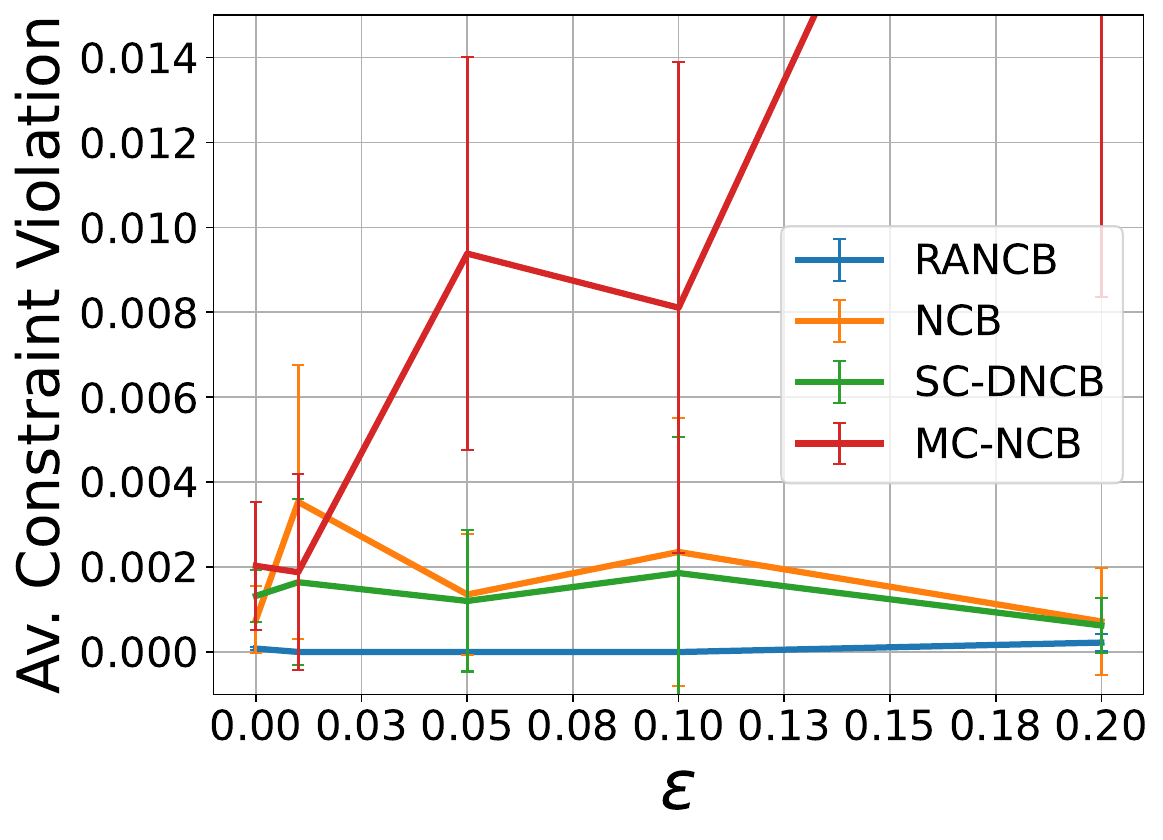}
\includegraphics[width=0.49\linewidth]{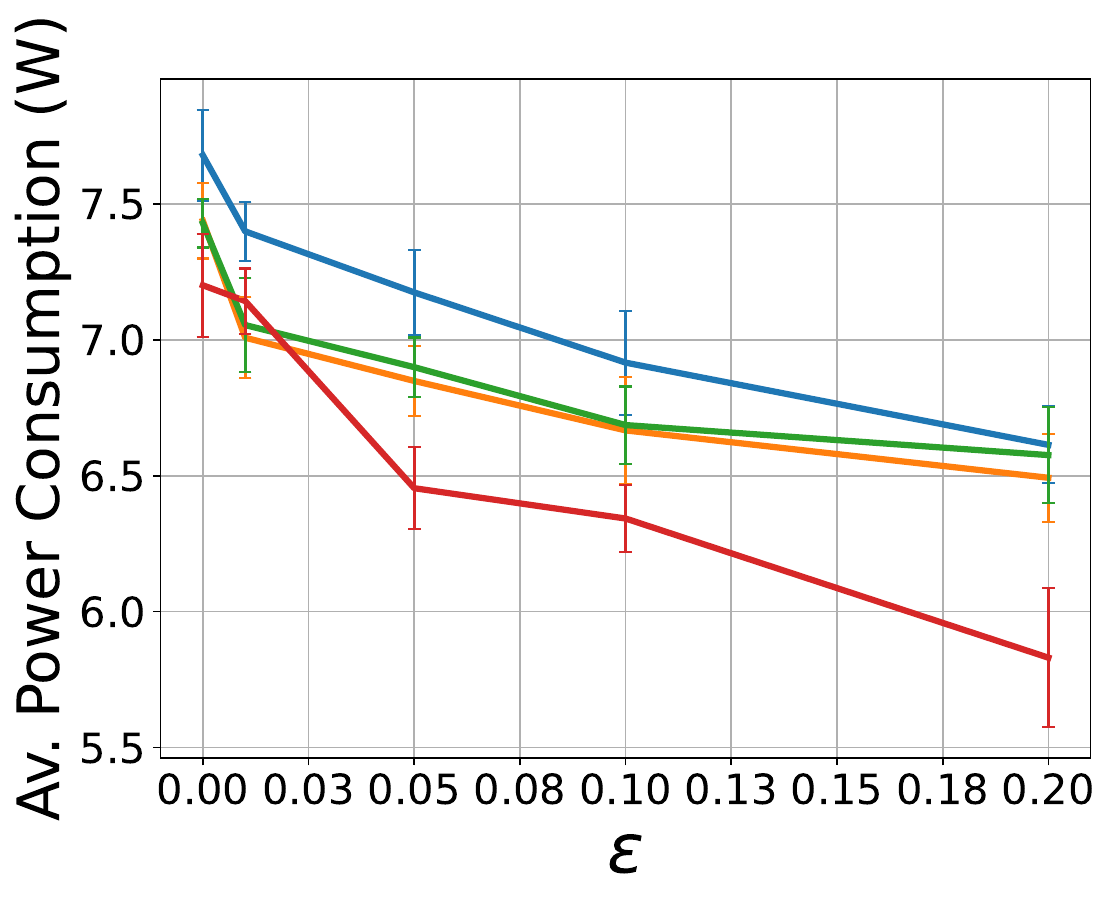}
\end{minipage}
\caption{\small{Performance evaluation in our wireless network experimental platform. Average constraint violation per step (left) and average power consumption in watts (right).}}
\label{fig:network_eval}
\end{figure}

The risk level (i.e., the tolerance to constraint violation) is determined by the specific application scenario. \citeauthor{risk_table} discuss different risk levels (reliability targets) for different scenarios (e.g., broadband communication services in cities vs. industry communication in factories). Moreover, network operators may want to reduce reliability in exchange for lower costs (energy consumption in this example) in some situations. For that reason, we evaluate various risk levels, showing our framework's flexibility to meet different application demands, balancing between risk and cost.

We consider 1500 iterations for training and 500 iterations for inference. Note that SafeOPT cannot be evaluated in this use case because, as shown in Fig~\ref{fig:times}, its inference time exceeds the system requirements (100 ms).
Fig.~\ref{fig:network_eval} depicts the mean system unreliability, i.e.,  $\frac{1}{T} \sum_{t=1}^{T} \zeta_t(s_t, a_t) - (1-\epsilon)$ (left) and the mean power consumption (right) in inference across all the solutions under study for various reliability targets $\epsilon$. Notably, \name{} consistently outperforms its benchmarks in terms of reliability, providing near-zero unreliability. Importantly, \name{}'s superior reliability comes at a low price in terms of power consumption, just 8.5\% higher than that of MC-NCB on average.

\section{Related Work}
There is a significant body of literature available on contextual bandit algorithms. Some works assume a structure in the reward function, e.g., a linear relationship between the contexts and the reward \cite{linucb, linear-cb-1, linear-cb-2}. Other works use neural networks (NN) to learn non-linear reward functions. For example, some use an NN to learn the embeddings of the actions and then apply Thompson Sampling on the last layer of the NN for exploration~\cite{neural_cb-1}. Others propose an NN-based algorithm with a UCB exploration~\cite{neural_cb-2}. \citeauthor{eenet} \shortcite{eenet} propose novel exploration strategies based on neural networks.
However, none of these works consider constraints in the problem formulation.

Other studies consider contextual bandits with budget constraints (bandits with knapsack) \cite{knapsack-cb-1, knapsack-cb-2}. However, the constraints in both of these papers are cumulative resources (budget) and deterministic, in contrast to the constraints in our problem that are stepwise and stochastic.
Others consider the linear contextual bandit problem with safety constraints \cite{lin-cb-constr-1, lin-cb-constr-2, TS-constr}. 
The goal of these works is to obtain at least a percentage of the performance of a baseline policy. Nevertheless, they do not consider the intrinsic random noise of the performance metrics and learn their mean value. Moreover, some of these~\cite{lin-cb-constr-1, lin-cb-constr-2} assume a structure in the reward function (linear), limiting their applicability to environments with non-linearities. 

The most closely related work to ours \cite{safeopt} proposes a Bayesian optimization algorithm called SafeOPT that, like in this work, handles constraints and noisy observations. 
This work generalizes the proposal of \citeauthor{sui2015safe} by considering multiple constraints and contexts.
In contrast to our approach, SafeOPT relies on Gaussian Processes (GPs) that are used to learn the objective and constraint functions. The GPs model the noise and the uncertainty in the estimation, which allows SafeOPT to compute a safe set of actions for each context. Besides, the use of GPs makes the algorithm very data-efficient.

However, SafeOpt has important drawbacks.
First, the use of GPs is computationally expensive. Specifically, the complexity scales as $O(n^3)$ with the number of data samples \cite{gp-book}. 
This hinders its application to settings requiring a large amount of data (e.g., environments with high dimensionality) and their deployment in computationally-constrained platforms.
Second, SafeOPT requires an initial set of actions that satisfy the constraints at the beginning of the training phase. As this set of actions can be highly dependent on the context, its computation can be very challenging, requiring some domain knowledge of the specific application, which limits the applicability of this solution. 
SafeOPT is objectively evaluated in comparison with our approach in the evaluation section.

Finally, in the Reinforcement Learning (RL) arena, there exist some works that do characterize the distribution of the value/Q function instead of its mean value \cite{c51,qr-dqn,iqn,mmd}. These approaches bring performance improvements even when only the mean of the distribution is used in the learning process. Moreover, some other works propose a distributional approach to optimize value-at-risk metrics in RL \cite{wcpg}.
However, to the best of our knowledge, these ideas have not been applied yet in the contextual bandit setting nor have they been used to design risk-aware decision-making algorithms in constrained environments.

\section{Conclusions}
This paper proposed a risk-aware decision-making framework for constrained contextual bandit problems.
The solution relies on an actor-multi-critic architecture, where the multiple critics characterize the distributions of the performance and constraint metrics, and a deterministic actor enables continuous control.
Our solution can adapt to different levels of risk to address the trade-off between constraint satisfaction and performance.
We evaluated our solution in a synthetic environment and a real-world mobile network testbed, showing its effectiveness.

\section{Acknowledgments}
The work was supported by the European Commission through Grants No. SNS-JU-101097083 (BeGREEN) and 101017109 (DAEMON). Additionally, it has been partially funded by the Spanish Ministry of Economic Affairs and Digital Transformation and the European Union – NextGeneration EU (Call UNICO I+D 5G 2021, ref. number TSI-063000-2021-3) and the CERCA Programme.

\bibliography{references}

\clearpage

\input{appendix}

\end{document}

%% file: appendix.tex
\appendix
\section*{Appendix}

\section{Evaluation of the hyperparameter $\lambda$}

We study the impact of $\lambda$ from eq.~\eqref{eq:ragg} on the performance of \name{}. This parameter weights the penalty incurred when a constraint is violated. We evaluate different values of $\lambda$ in the synthetic environment from eq.~\eqref{eq:syn_env}.

\begin{figure}[t!] 
\centering
\begin{minipage}{\linewidth}
\centering
\includegraphics[width=0.49\linewidth]{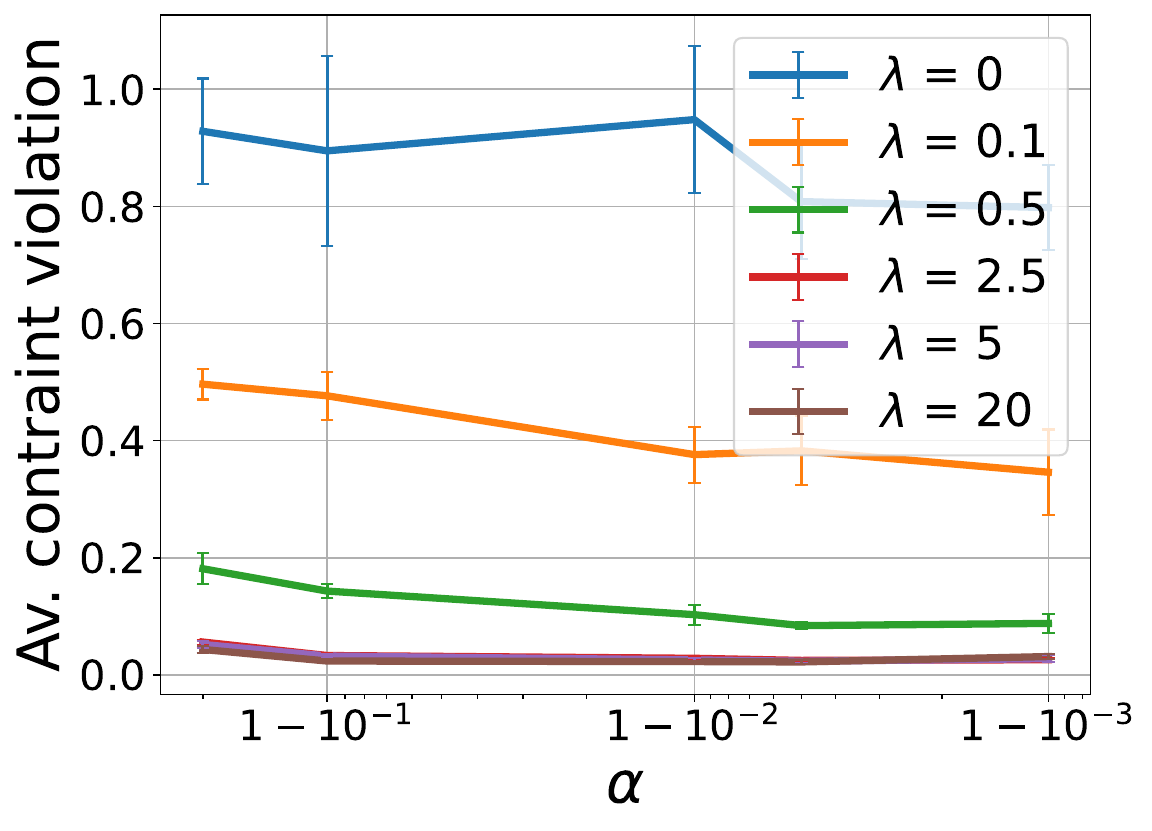}
\includegraphics[width=0.49\linewidth]{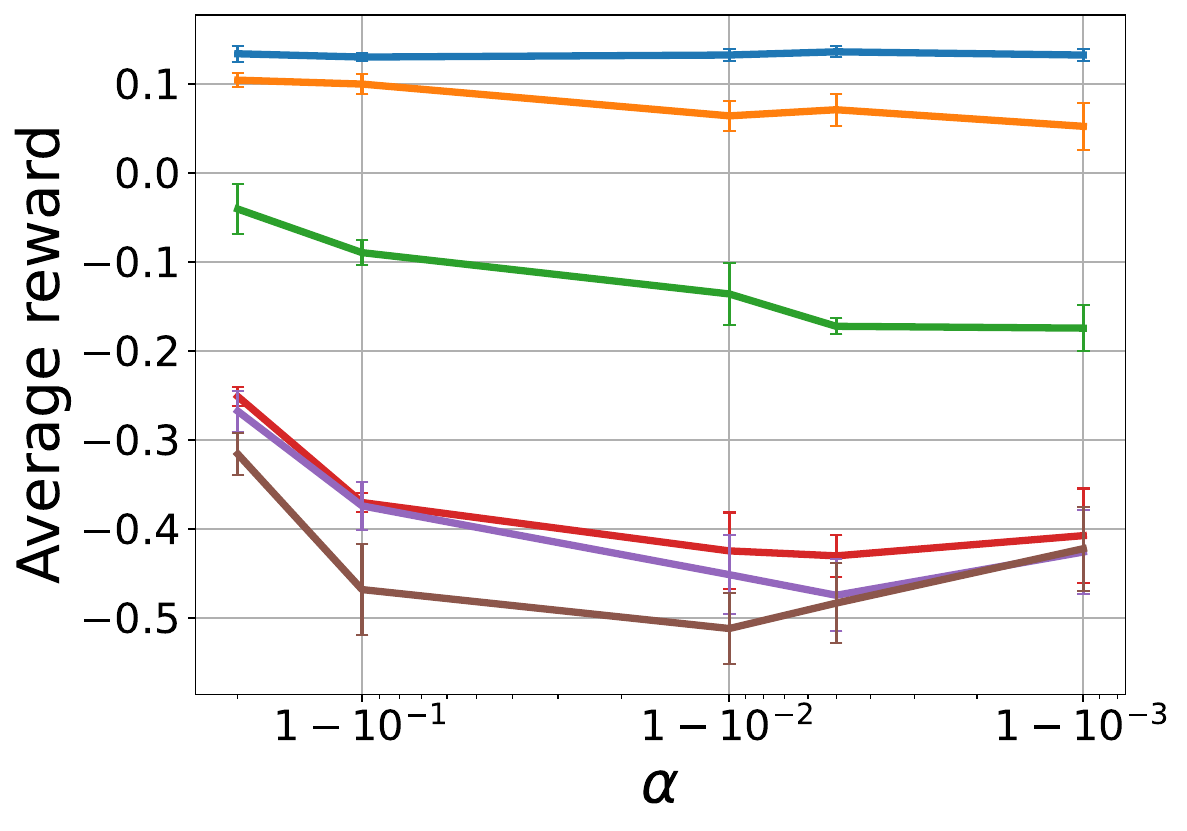}
\end{minipage}
\caption{\small{Average constraints violation and the average reward for different values of $\lambda$ in the synthetic environment.}}
\label{fig:lambda_eval}
\end{figure}

Fig.~\ref{fig:lambda_eval} shows the average constraints violation and the average reward for different values of $\lambda$. We consider the same parameters as in the evaluation section of the paper and $\sigma_{\text{env}} = 0.2$. We observe that for values of $\lambda$ greater than $2.5$, there is no improvement in the constraint violation. Conversely, the reward decreases for larger values of $\lambda$ as the resolution of $R^{agg}(\cdot)$ worsens. 
We observe that this result is consistent across different scenarios due to the normalization of the reward and cost functions.
Thus, we use $\lambda = 2.5$ for all the evaluations in this work.

\section{Evaluation of Dimensionality}
In this appendix, we evaluate the impact of the dimensionality of the contexts and actions on the learning performance. For that purpose, we define a new environment as follows:
\begin{gather}
r_t(s_t,a_t) = \sum_{i=1}^{D} s_t^{(i)} \cdot (a^{(i)}_{t})^i \nonumber \\
c_t(s_t,a_t) = \sum_{i=1}^{D} (-1)^{i} \cdot s_t^{(i)} \cdot (a^{(i)}_{t})^i, \nonumber
\end{gather}
where $D$ indicates the dimensionality of both the context $s_t$ and the continuous action vector $a_t$. Note that both the reward and constraint functions are polynomials of degree $D$. We evaluate $D = \{2,\ldots,25\}$ considering the parameter configuration described in the evaluation section of the paper and $\sigma_{\text{env}} = 0.2$.

\begin{figure}[t!] 
\centering
\begin{minipage}{\linewidth}
\centering
\includegraphics[width=0.49\linewidth]{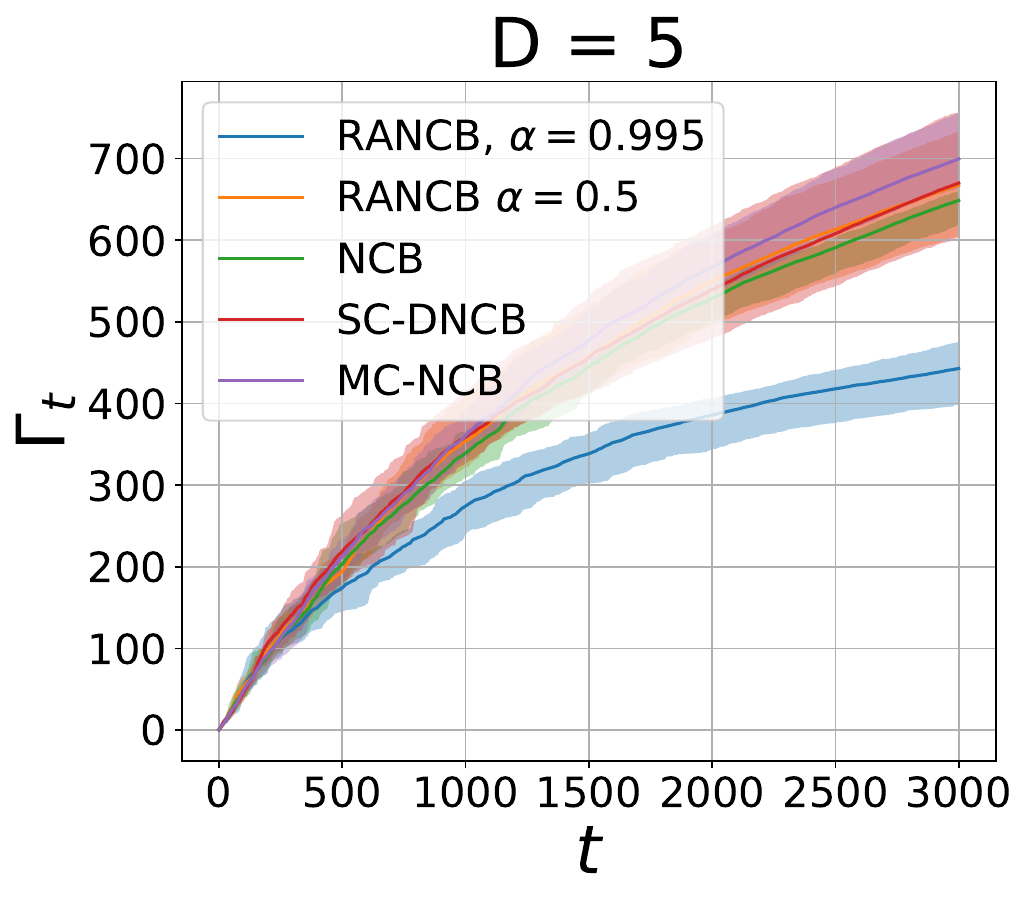}
\includegraphics[width=0.49\linewidth]{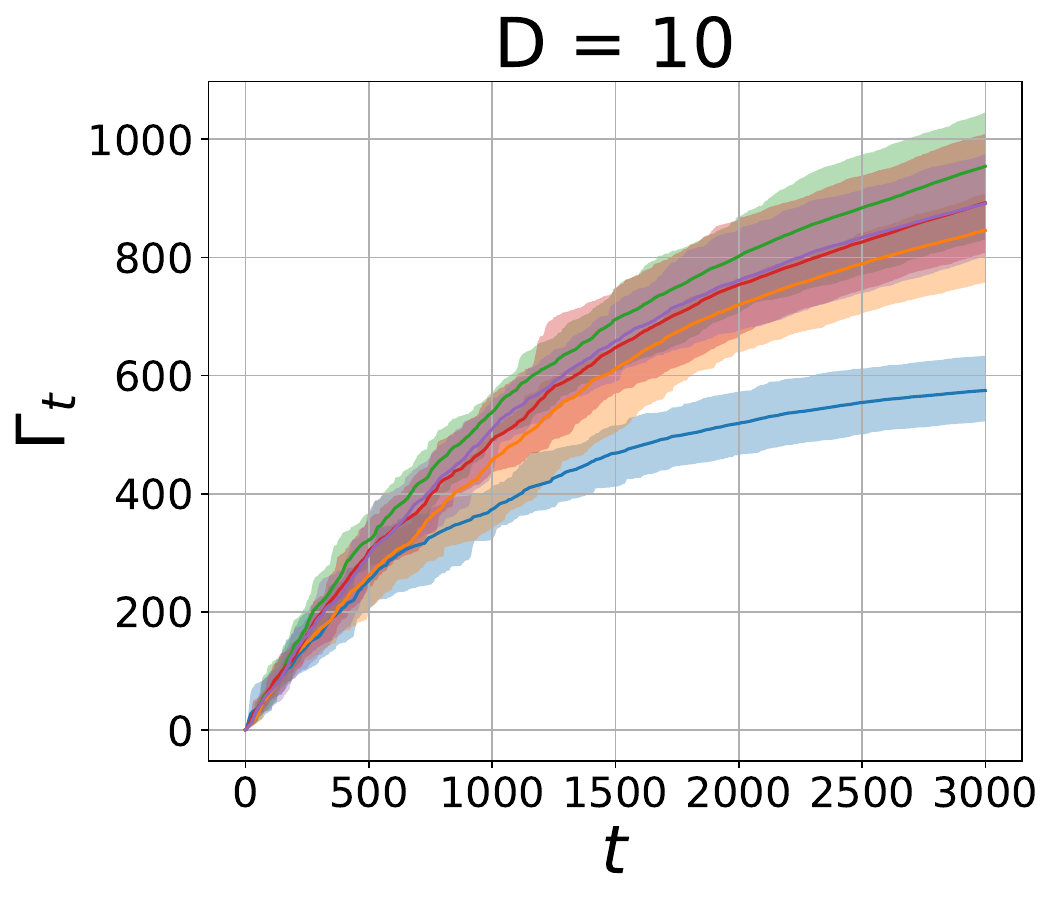}
\includegraphics[width=0.49\linewidth]{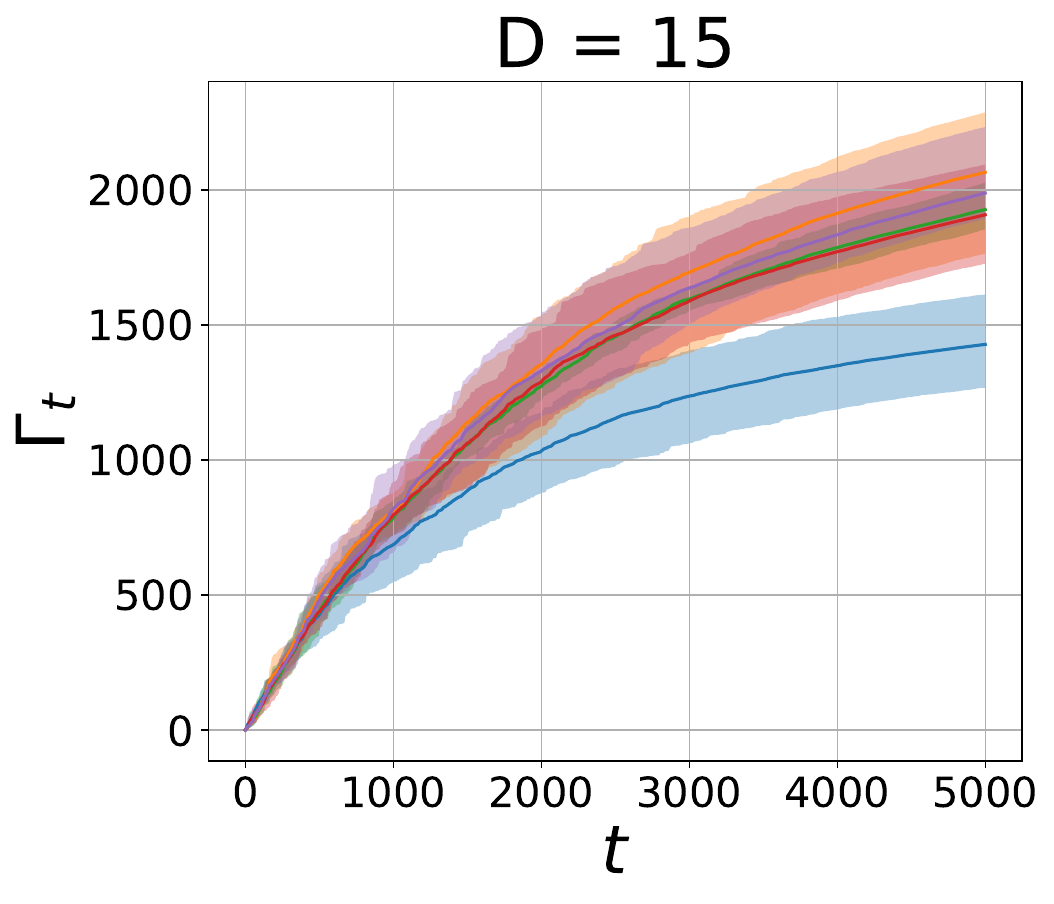}
\includegraphics[width=0.49\linewidth]{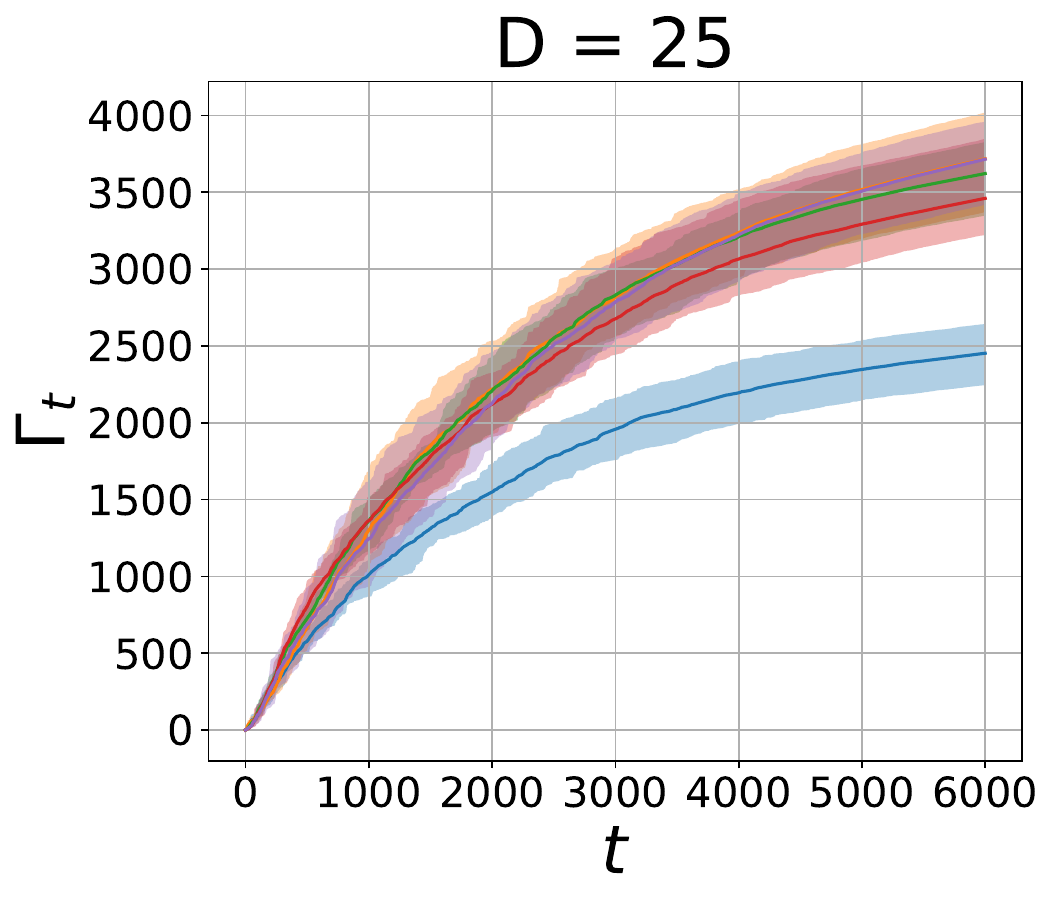}
\end{minipage}
\caption{\small{Accumulated constraint violation during training phase for different dimensionalities of the contexts and action.}}
\label{fig:dim_eval}
\end{figure}

Fig.~\ref{fig:dim_eval} shows that the gains of our proposal in terms of constraint satisfaction during the training phase further increase with higher dimensionality compared to Fig.~\ref{fig:syn_train}. Note that, as expected, more training steps are needed to reach convergence with higher dimensionality.

\section{Resource Assignment in Mobile Networks}
In this appendix, we detail the computing resource assignment problem in mobile networks addressed in this work.

The Transmission Time Interval (TTI) is the time interval, typically 1~ms or lower, that a base station (BS) has to schedule wireless transmissions encoding data, known as Transport Blocks (TB). We index each TTI with $k =1, 2, \ldots$. Every TTI $k$, the BS receives from its users a set of encoded TBs denoted by $\mathcal{B}_k$ that must be decoded by a processing unit  $p \in \{\text{CPU}, \text{HA}\}$.
Every TB $b_i \in \mathcal{B}^b_k$ is characterized by its signal-to-noise power relationship ($b_i^c$), the modulation and coding scheme ($b_i^m$), and the amount of data bits it carries ($b_i^l$). 
We let $E_{b_i}(b_i^c, b_i^m, b_i^l, p)$ denote the energy consumed by a processing unit $p$ to decode a TB. Similarly, we define the TB processing time as $P_{b_i}(b_i^c, b_i^m, b_i^l, p)$. 

In our experiments, we observed that, for a given TB and processing unit, both the energy and the processing time are non-deterministic. 
When a TB $b_i$ arrives at a processing unit $p$ to be decoded, if the processing unit is already busy, $b_i$ is stored in a FIFO queue until $p$ becomes empty. Following the specification of the relevant standardization bodies, every TB that spends more than a time $P_{\text{max}}$ in the system without being decoded is discarded, incurring data loss. A TB can be discarded while waiting in the FIFO queue or while being processed. In the latter case, in addition to the data loss, there is an additional energy waste associated with the processing of outdated data.

As mentioned in the paper, in novel Open RAN systems, control decisions may be taken by third-party applications at near-real-time decision periods of $\sim$100~ms.
We denote by $\mathcal{B}_t = \{\mathcal{B}_k \mid k \in \mathcal{K}_t\}$ the set of TBs generated during a decision period $t$, where $\mathcal{K}_t$ indicates the set of TTIs belonging to $t$.
Then, the energy $E_t(\cdot)$ and reliability $\zeta_t(\cdot)$ defined in eq.~\eqref{eq:network_problem} are computed based on all the TBs in $\mathcal{B}_t$ with varying parameters ($b_i^c$, $b_i^m$, and $b_i^l$), and considering that some of them are queued, discarded, etc. The complexity of this system makes predicting energy consumption and reliability performance very difficult.

The state $s_t$ in eq.~\eqref{eq:network_problem} shall characterize the traffic patterns within a period $t$ in terms of the TBs generated by the users. Particularly, we define $s_t = \Phi(\mathcal{B}_t, D)$ as the 3-dimensional histogram of the TB’s features (signal quality, modulation, and coding scheme, and TB size), where $D$ is the number of bins of this histogram in each dimension. In our experiments, we use $D = 5$ bins.

Finally, the offloading strategy used in eq.~\eqref{eq:network_problem} is denoted by $a_t \in [0, 1]$. A straightforward approach is that $a_t$ indicates the portion of traffic (in terms of the number of TB) assigned to each processing unit. 
However, based on the insights from our experiments, we devised a better alternative. 
We observed that GPUs can process larger TBs much faster than CPUs; in fact, CPUs are unable to process some large TBs within their deadline. We also observed in our experiments that CPUs consume less energy than the GPU to process TBs. Motivated by these observations, we define $a_t$ as the normalized \emph{bit threshold}. Thus, when $b_i^l$ is larger than the bit threshold, $b_i$ is processed in the HA and otherwise processed in the CPU.

\section{Open RAN experimental platform}

\begin{figure}[t!] 
\centering
\includegraphics[width=\linewidth]{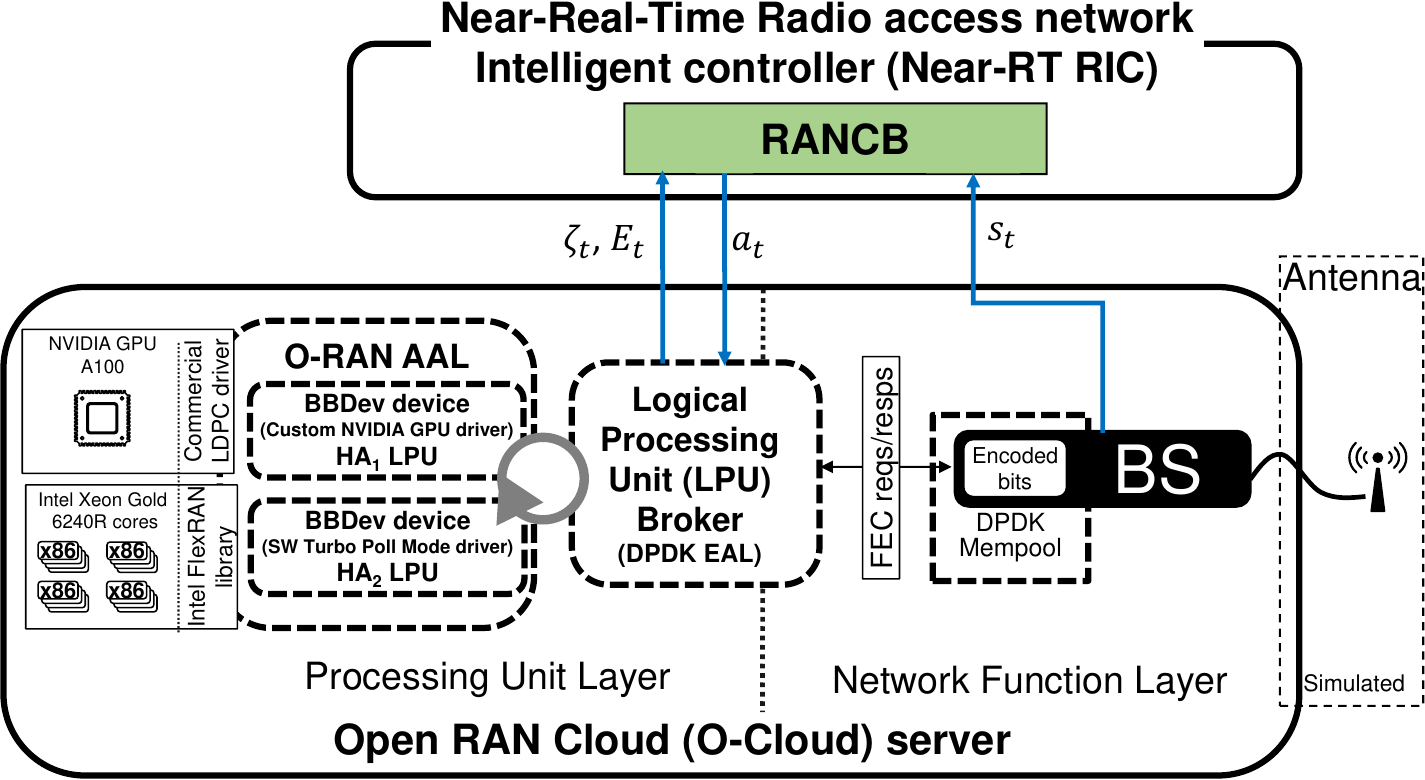}
\caption{\small{Open RAN (O-RAN) Cloud server for 5G BSs.}}
\label{fig:prototype}
\end{figure}

Our experimental platform, depicted in Fig.~\ref{fig:prototype}, complies with the technical specification of the novel Open RAN (O-RAN) standards for 5G base stations \cite{oran-magazine}. More specifically, our platform comprises a general-purpose server with an Intel Xeon Gold 6240R CPU with 16 cores dedicated to signal processing tasks (CPU processing unit) and an NVIDIA GPU V100 used as a hardware accelerator (HA processing unit).
Within the O-Cloud, an Acceleration Abstraction Layer (AAL)~\cite{oran-aal-spec} interfaces between the BS software and processing units for signal processing (CPU or HA). The AAL provides a Logical Processing Unit (LPU), a driver to manage computing resources for signal processing, for each processing unit. To this end, we used Intel DPDK BBDev\footnote{\url{https://doc.dpdk.org/guides/prog_guide/bbdev.html}}, a software library from Intel (FlexRAN~\cite{flexran}) to implement the CPU LPU, and proprietary driver to implement the GPU LPU.

\begin{figure}[t!] 
\centering
\begin{minipage}{\linewidth}
\centering
\includegraphics[width=0.49\linewidth]{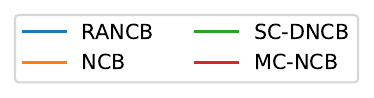}
\begin{minipage}{\linewidth}
\centering
\includegraphics[width=0.49\linewidth]{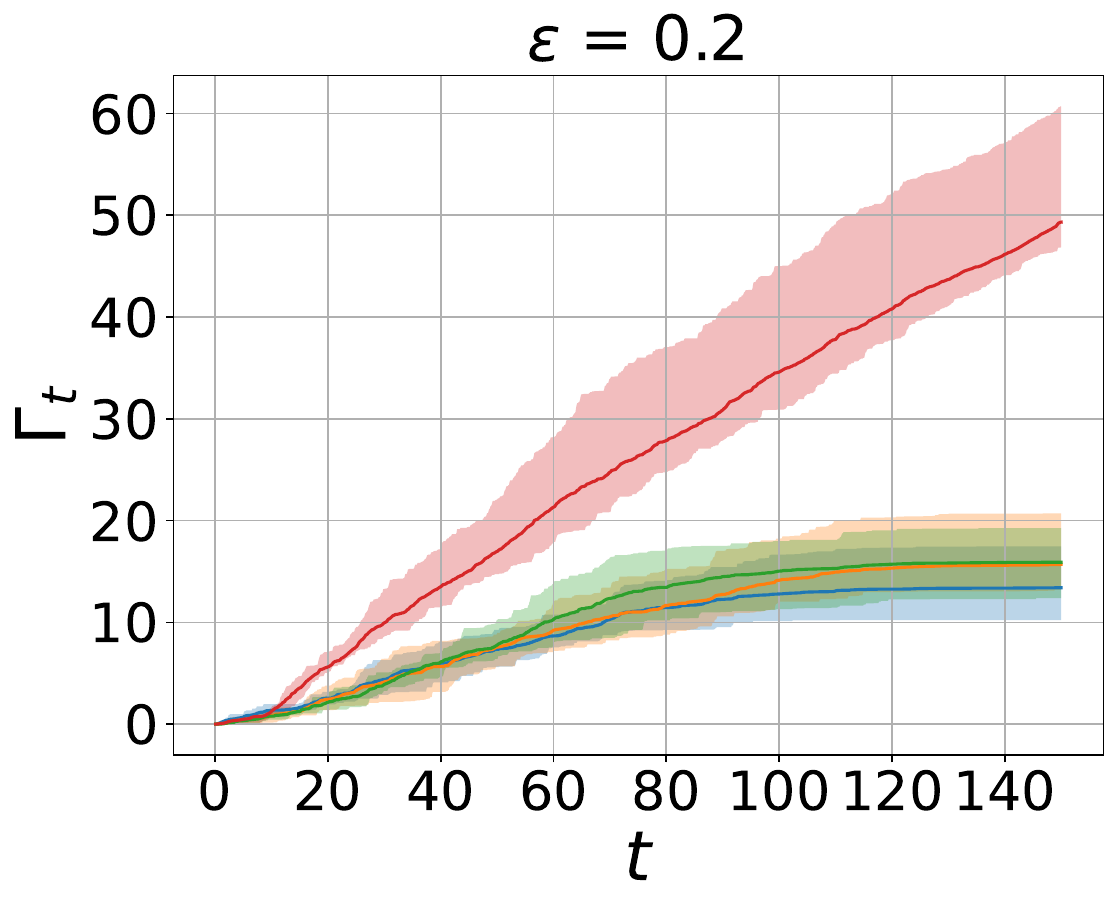}
\includegraphics[width=0.49\linewidth]{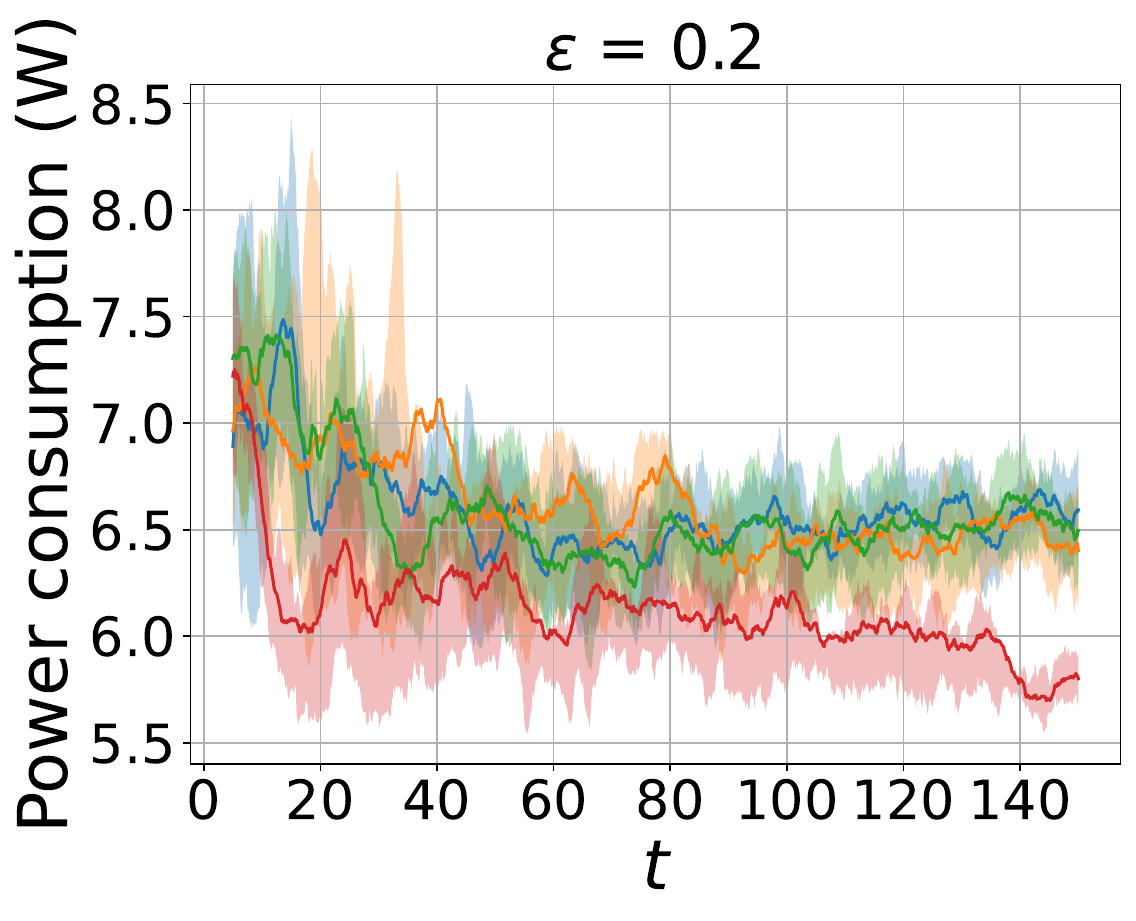}
\includegraphics[width=0.49\linewidth]{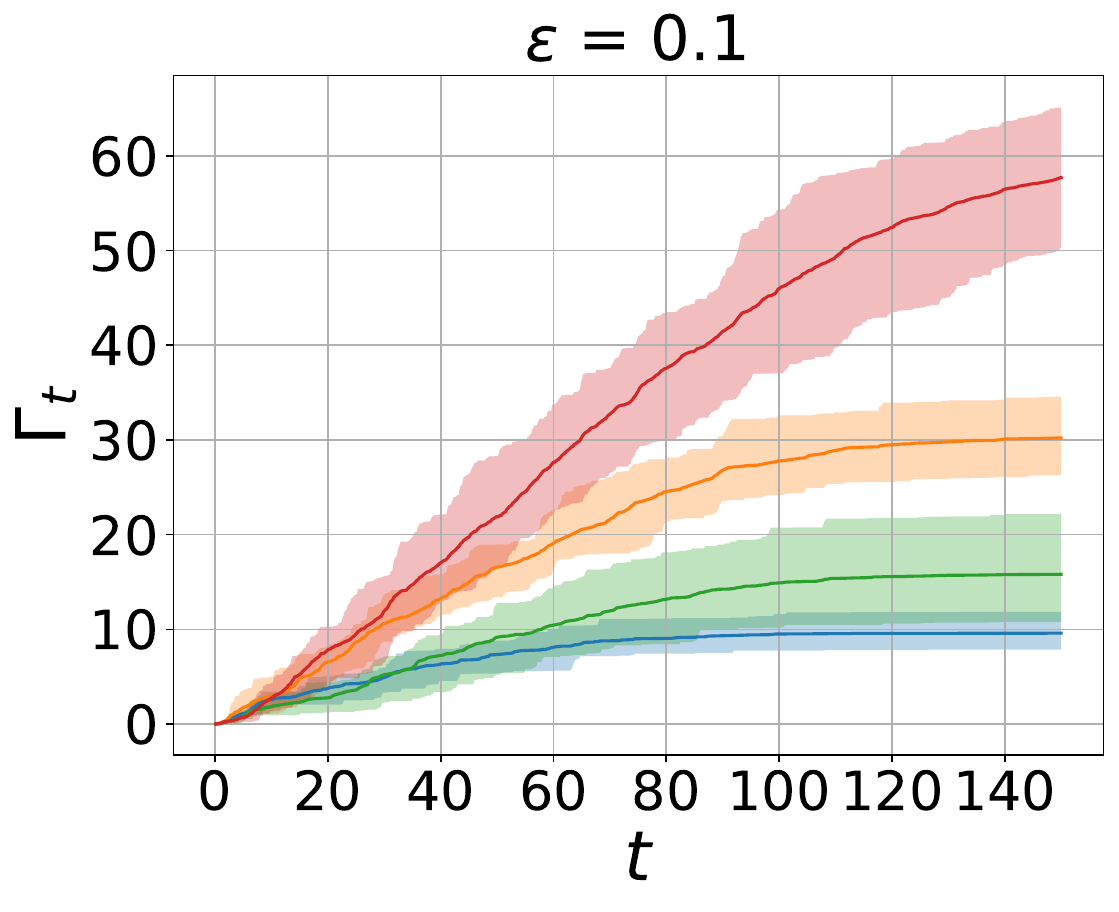}
\includegraphics[width=0.49\linewidth]{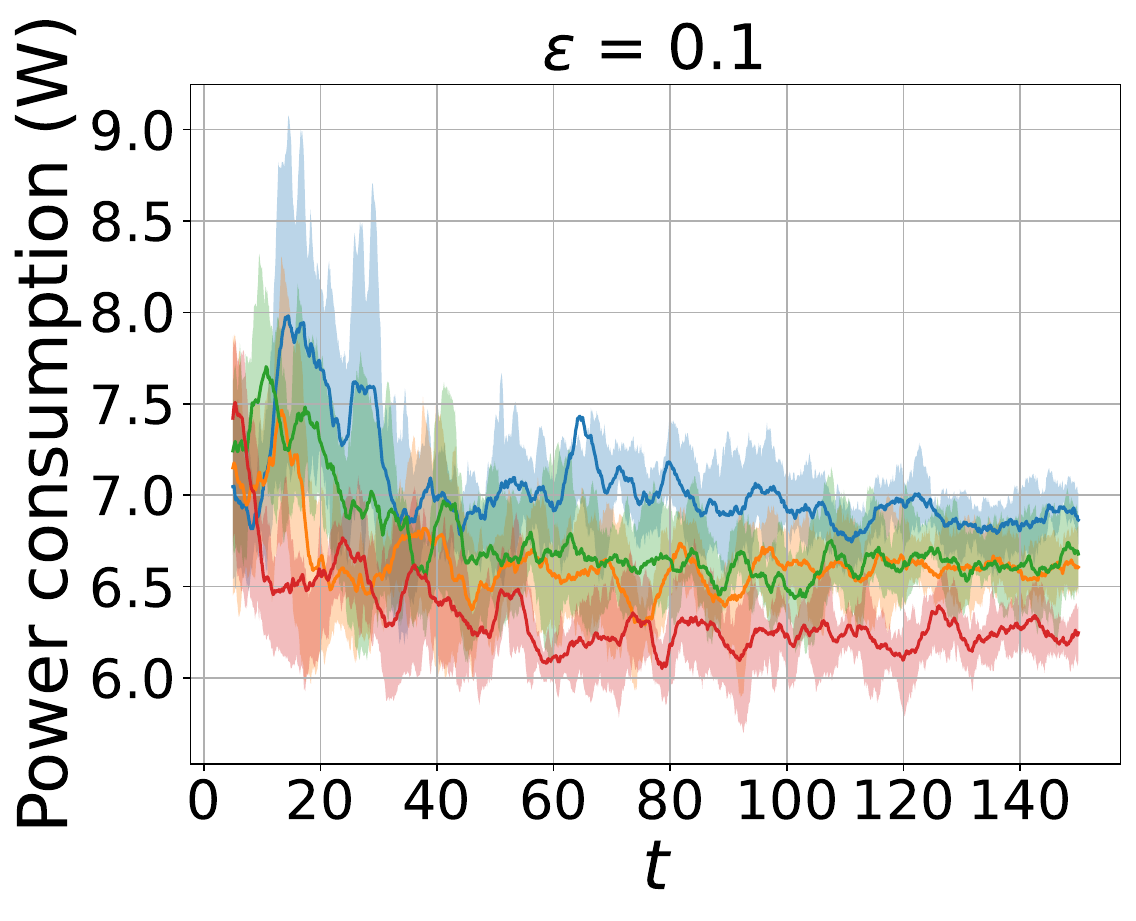}
\includegraphics[width=0.49\linewidth]{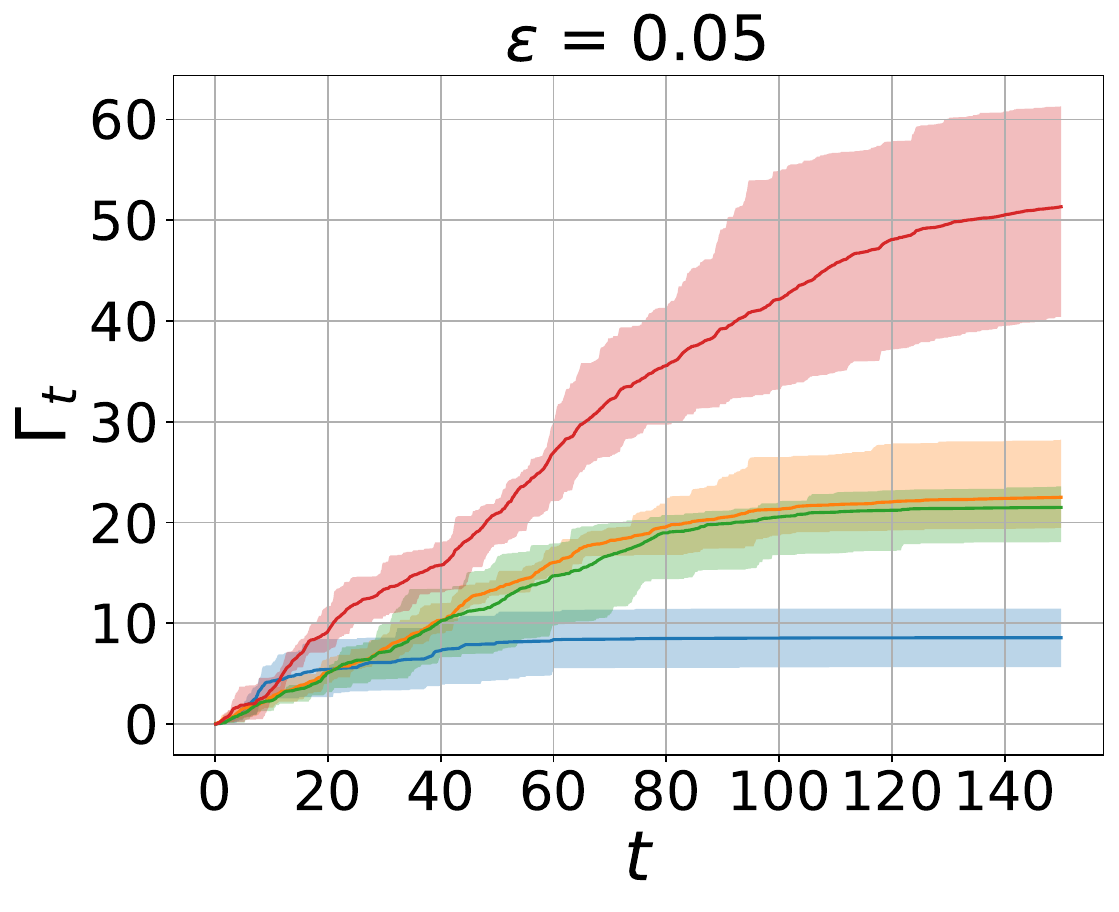}
\includegraphics[width=0.49\linewidth]{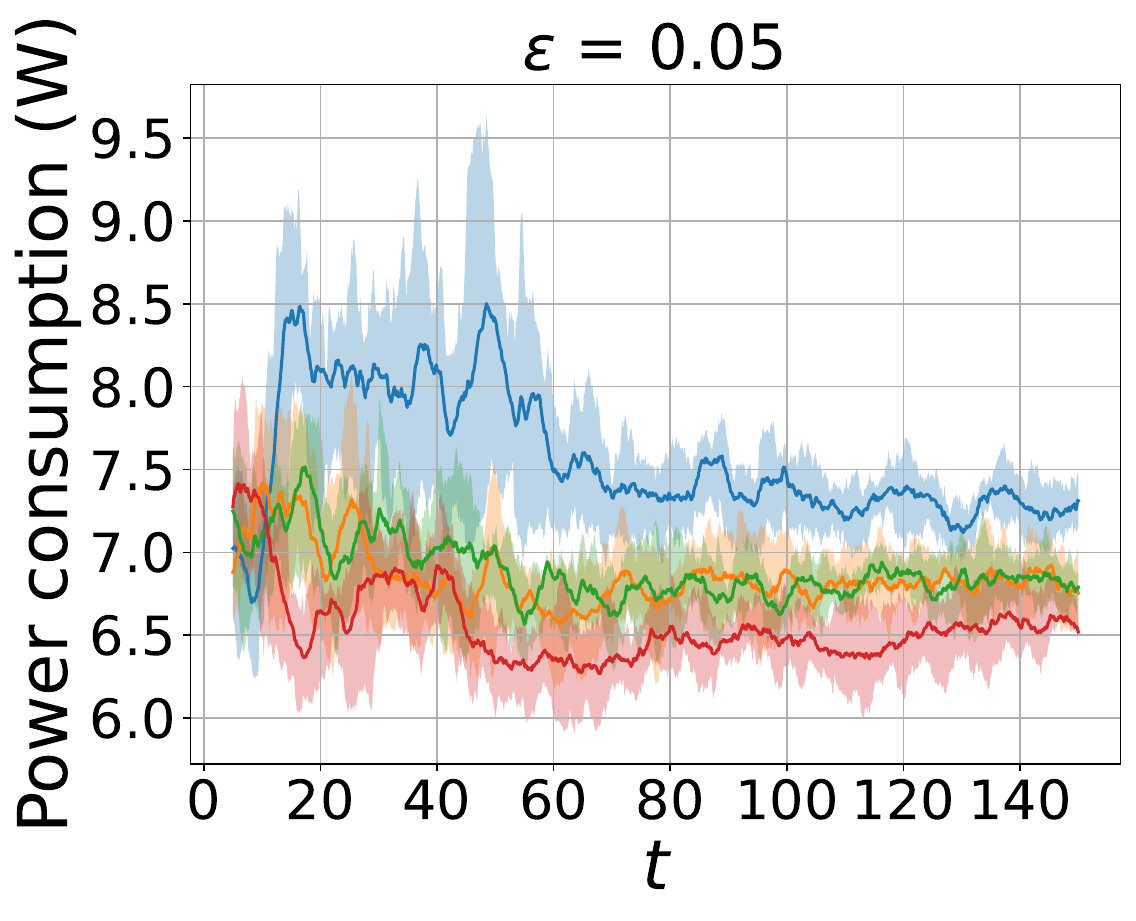}
\includegraphics[width=0.49\linewidth]{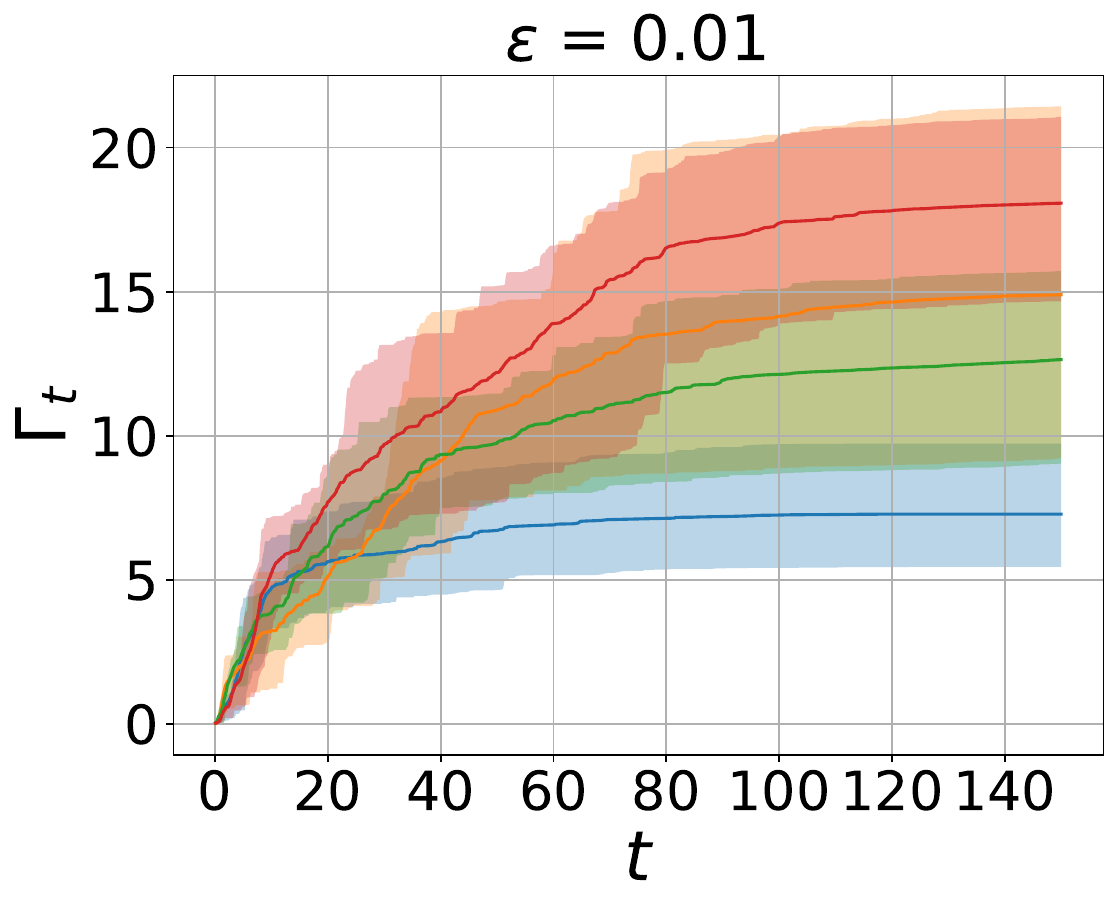}
\includegraphics[width=0.49\linewidth]{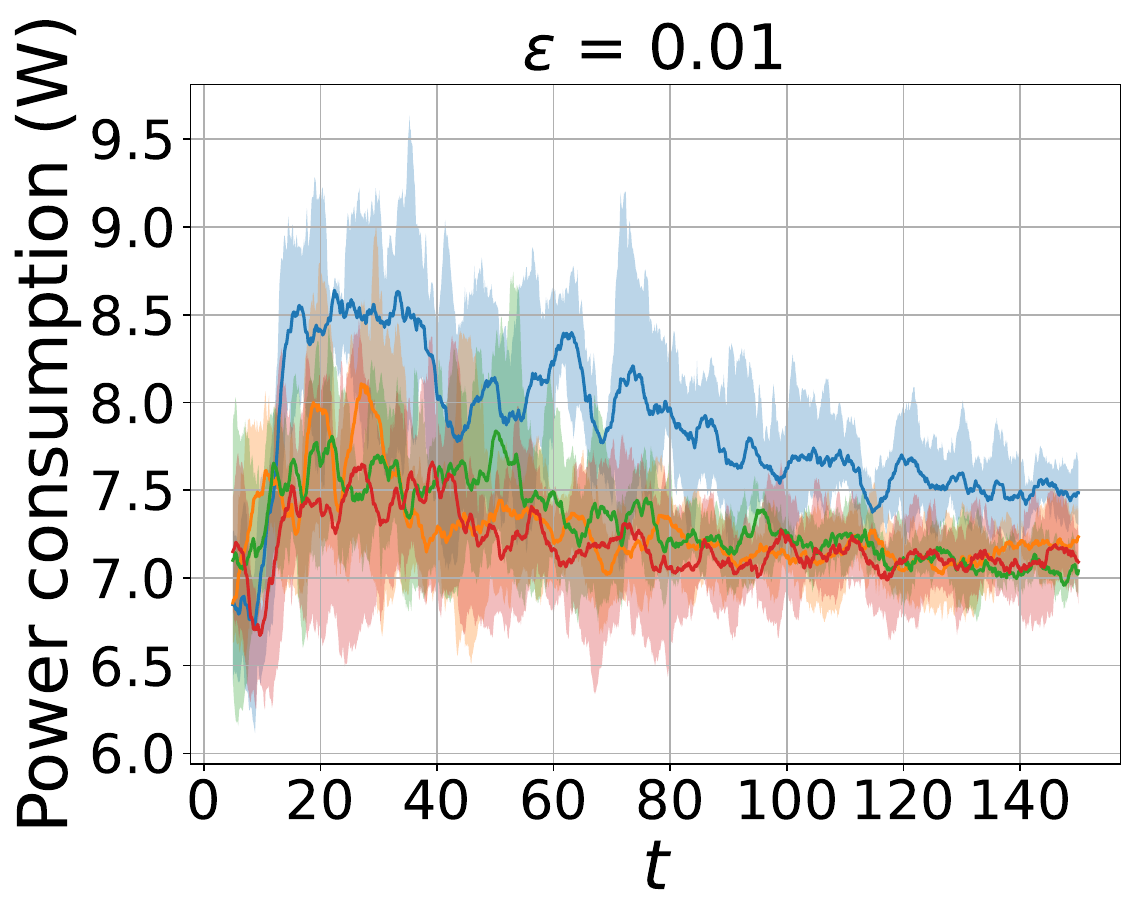}
\includegraphics[width=0.49\linewidth]{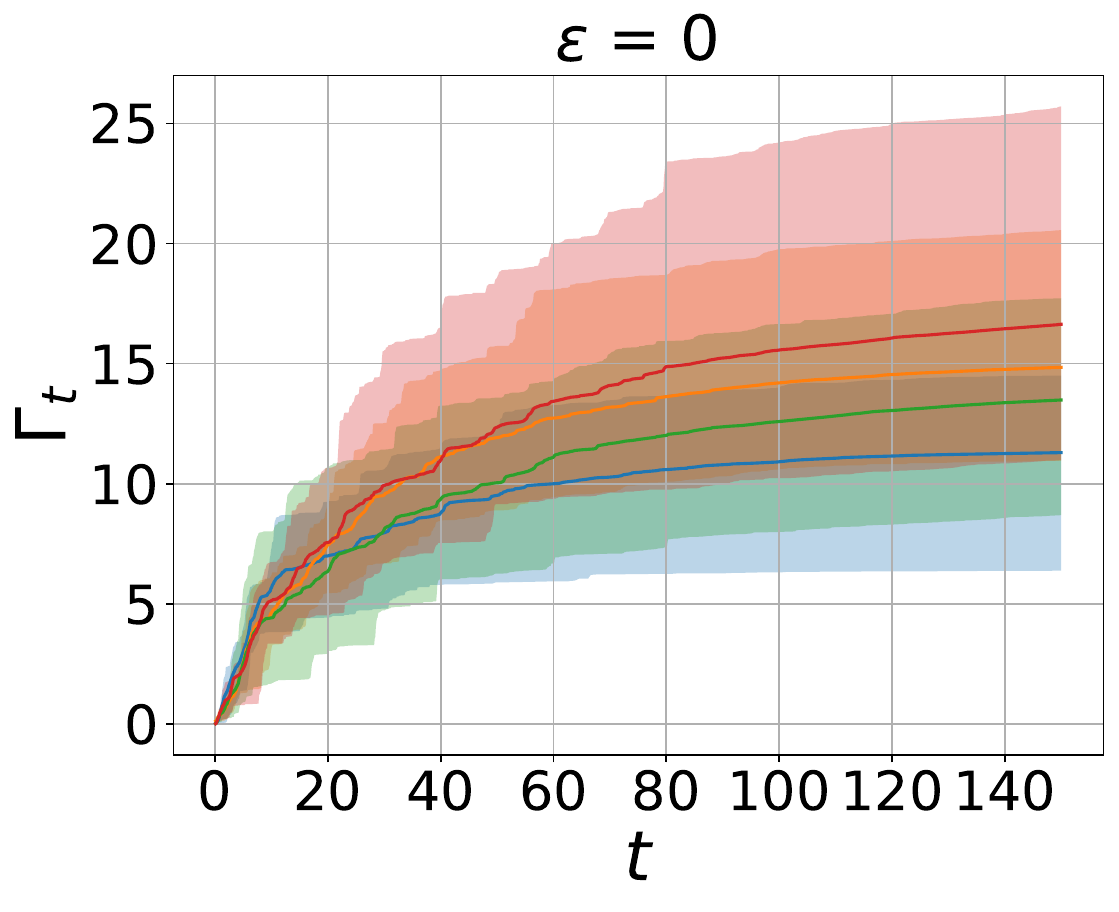}
\includegraphics[width=0.49\linewidth]{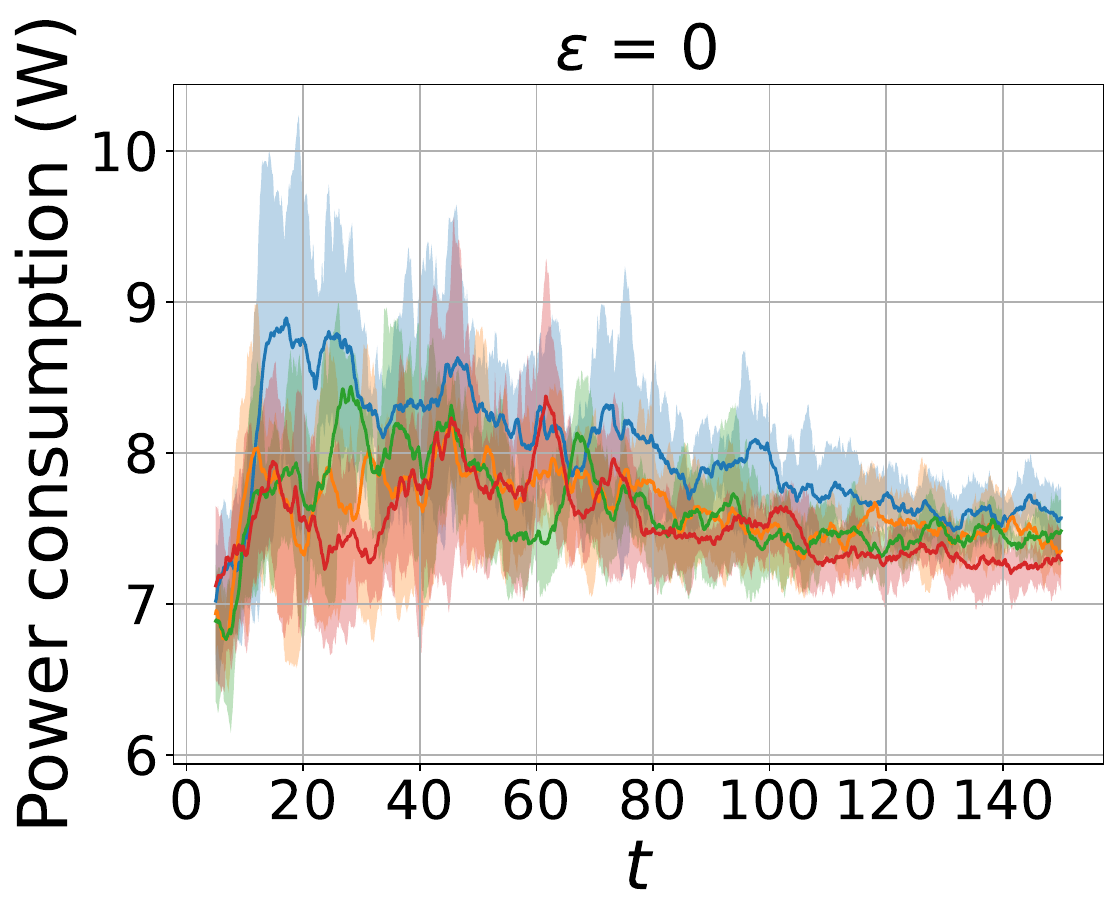}
\end{minipage}
\end{minipage}
\caption{\small{Training phase 
resource assignment problem in mobile networks
accumulated constraint violation (first column) and power consumption (second column) for different reliability targets $\epsilon$ (rows).}}
\label{fig:network_training}
\end{figure}

We emulate 5G signals by using real traces collected from real BSs in Madrid, Spain, during May 2023 using an open-source tool called Falcon~\cite{falcon}. In these traces, we collect information related to the amount of data sent by each user, their signal quality, etc. We do not collect personal information nor actual user data since, per standard, it is encrypted. 
We then emulate realistic patterns of 5G signals by encoding and modulating TBs with the same characteristics of size, signal quality, etc. as in our traces. The resulting TBs are sent to the BS as shown in Fig.~\ref{fig:networking_scheme}. We emulate 6 concurrent users connected to the BS.

\name{} and the benchmarks run as applications within O-RAN's Near-Real Time Radio access network Intelligent Controller (Near-RT RIC), which controls the behavior of BSs in timescales of $\sim$100~ms. In our case, the Near-RT RIC is deployed in a general-purpose server with an Intel i7-11700 @ 2.5GHz and 15Gb or RAM.
On the one hand, an LPU broker implements the aforementioned threshold rule. Upon the arrival of each TB, the LPU broker directs the TB to the corresponding LPU according to such threshold and reports back performance metrics to the Near-RT RIC. On the other hand, the BS provides context information to the Near-RT RIC through an O-RAN interface called O2. 


\section{Learning curves mobile networks}
This appendix provides the training phase results of the use case of resource assignment in mobile networks. Fig.~\ref{fig:network_training} shows the evolution of the accumulated constraint violation in the first column and the evolution of power consumption in the second one. Each row in Fig.~\ref{fig:network_training} shows a reliability target $\epsilon$ corresponding to all the points in the x-axis of Fig.~\ref{fig:network_eval}. In all the cases our proposal incurs the minimum constraint violation and pays a cost in terms of power, which illustrates the trade-off between reliability and energy consumption. 
%